# SAP Speaks PDDL: Exploiting a Software-Engineering Model for Planning in Business Process Management

**Jörg Hoffmann**                                    HOFFMANN@CS.UNI-SAARLAND.DE
*Saarland University, Saarbrücken, Germany*

**Ingo Weber**                                       INGO.WEBER@NICTA.COM.AU
*NICTA, Sydney, Australia*

**Frank Michael Kraft**              FRANK.MICHAEL.KRAFT@BPMNFORUM.NET
*bpmnforum.net, Germany*

## Abstract

Planning is concerned with the automated solution of action sequencing problems described in declarative languages giving the action preconditions and effects. One important application area for such technology is the creation of new processes in Business Process Management (BPM), which is essential in an ever more dynamic business environment. A major obstacle for the application of Planning in this area lies in the modeling. Obtaining a suitable model to plan with – ideally a description in PDDL, the most commonly used planning language – is often prohibitively complicated and/or costly. Our core observation in this work is that this problem can be ameliorated by leveraging synergies with model-based software development. Our application at SAP, one of the leading vendors of enterprise software, demonstrates that even one-to-one model re-use is possible.

The model in question is called Status and Action Management (SAM). It describes the behavior of Business Objects (BO), i.e., large-scale data structures, at a level of abstraction corresponding to the language of business experts. SAM covers more than 400 kinds of BOs, each of which is described in terms of a set of status variables and how their values are required for, and affected by, processing steps (actions) that are atomic from a business perspective. SAM was developed by SAP as part of a major model-based software engineering effort. We show herein that one can use this same model for planning, thus obtaining a BPM planning application that incurs no modeling overhead at all.

We compile SAM into a variant of PDDL, and adapt an off-the-shelf planner to solve this kind of problem. Thanks to the resulting technology, business experts may create new processes simply by specifying the desired behavior in terms of status variable value changes: effectively, by describing the process in their own language.

## 1. Introduction

Business processes are workflows controlling the flow of activities within and between enterprises (Aalst, 1997). Business process management (BPM) is concerned, amongst other things, with the maintenance of these processes. To minimize time-to-market in an ever more dynamic business environment, it is essential to be able to quickly create new processes. Doing so involves selecting and arranging suitable IT transactions from huge in-frastructures. That is a very difficult and costly task. Our application supports this task within the software framework of SAP[1], one of the leading vendors of enterprise software.

A well-known idea in this context, discussed for example by Jonathan, Moore, Stader, Macintosh, and Chung (1999), Biundo, Aylett, Beetz, Borrajo, Cesta, Grant, McCluskey, Milani, and Verfaillie (2003), and Rodriguez-Moreno, Borrajo, Cesta, and Oddi (2007), is to use technology from the field of *planning*. This is a long-standing sub-area of AI, that allows the user to describe the problem to be solved in a declarative language. In a nutshell, planning problems come in the form of an initial state, a goal, and a set of actions, all formulated relative to a set of (typically Boolean or at least finite-domain) state variables. A solution (or "plan") is a schedule of actions transforming the initial state into a state that satisfies the goal. The planning technology solves (in principle) any problem described in that language. By far the most wide-spread planning language is the *planning domain definition language (PDDL)* (McDermott, Ghallab, Howe, Knoblock, Ram, Veloso, Weld, & Wilkins, 1998).[2]

The idea in the BPM context is to annotate each IT transaction with a planning-like description formalizing it as an action. This enables planning systems to compose (parts or approximations of) the desired processes fully automatically, i.e., based on minimal user input specifying from where the process will start (initial state), and what it should achieve (goal). Very closely related ideas have been explored under the name *semantic web service composition* in the context of the Semantic Web community (e.g., Narayanan & McIlraith, 2002; Agarwal, Chafle, Dasgupta, Karnik, Kumar, Mittal, & Srivastava, 2005; Sirin, Parsia, & Hendler, 2006; Meyer & Weske, 2006).

Runtime performance is important in such an application. Typically, the user – a business expert wishing to create a new process – will be waiting online for the planning outcome. However the most mission-critical question, discussed for example by Kambhampati (2007) and Rodriguez-Moreno et al. (2007), is: *How to get the planning model?* To be useful, the model needs to capture the relevant properties of a huge IT infrastructure, at a level of abstraction that is high-level enough to be usable for business experts, and at the same time precise enough to be relevant at IT level. Designing such a model is so costly that one will need good arguments indeed to persuade a manager to embark on that endeavor.

In the present work, we demonstrate that this problem can be ameliorated by leveraging synergies with model-based software development, thus reducing the additional modeling overhead caused by planning. In fact, we show that one can – at least in our particular application – *re-use exactly, one-to-one, models that were built for the purpose of software engineering, and thus reduce the modeling overhead to zero.*

It has previously been noted, for example by Turner and McCluskey (1994) and Kitchin, McCluskey, and West (2005), that planning languages have commonalities with software specification languages such as B (Schneider, 2001) and OCL (Object Management Group, 2006). Now, typically such specification languages are mathematically oriented to describe low-level properties of programs. This stands in contrast with the more abstract models needed to work with business experts. But that is not always so.

---

1. http://www.sap.com
2. There are many variants of planning, and of PDDL. All share concepts similar to the short description we just stated. However, that description corresponds best to "classical planning", where (amongst other things) there is no uncertainty about the action effects. We will discuss some details in Section 2.1. Throughout the paper, unless we refer to one of the particular planning formalisms defined in here, we use the term "planning" in a general sense not targeting any particular variant.

As part of a major effort developing a flexible service-oriented (Krafzig, Banke, & Slama, 2005; Bell, 2008) IT infrastructure, called *SAP Business ByDesign*, SAP has developed a model called *Status and Action Management (SAM)*. SAM describes how "status variables" of Business Objects (BO) change their values when "actions" – IT transactions affecting the BOs – are executed. BOs in full detail are vastly complex, containing 1000s of data fields and numerous technical-level transactions. SAM captures the more abstract business perspective, in terms of a smaller number of user-level actions (like "submit" or "reject"), whose behavior is described using preconditions and effects on high-level status properties (like "submitted" or "rejected"). In this way, SAM corresponds to the language of business users, and is in very close correspondence with common planning languages. SAM is extensive, covering 404 kinds of BOs with 2418 transactions. The model creation in itself constitutes a work effort spanning several years, involving, amongst other things, dedicated modeling environments and educational training for modelers.

SAM was originally designed for the purpose of model-driven software development, to facilitate the design of the Business ByDesign infrastructure, and changes thereunto during its initial development and afterwards. Business ByDesign covers the needs of a great breadth of different SAP customer businesses, and is flexibly configurable for these customers. That configuration involves, amongst other things, the design of customer-specific processes, appropriately combining the functionalities provided. Describing the properties of individual processing steps, rather than supplying each BO with a standard life-cycle workflow, SAM is well-suited to support this flexibility. However, the business users designing the processes are typically not familiar with the details of the infrastructure. Using SAM for planning, we obtain technology that alleviates this problem. As its output, our technology delivers a first version of the desired process, with the relevant IT transactions and a suitable control-flow. As its input, the technology requires business users only to specify the desired status changes – in their own language.

The intended meaning of SAM is, to a large extent, the same as in common planning frameworks. There are some subtleties in the treatment of non-deterministic actions. One problem is that many of the non-deterministic actions modeled in SAM have "bad" outcomes that preclude successful processing of the respective business object (example: "BO data inconsistent"). That problem is aggravated by the fact that, in SAM's "non-determinism", repeated executions of the same action are not independent (example: "check BO consistency"). We discuss this in detail, and derive a suitable planning formalism. We compile SAM into PDDL, thus creating as a side-effect of our work a new planning benchmark. An anonymized PDDL version of SAM is publicly available.

On the algorithmic side, we show that minimal changes to an off-the-shelf planner suffice to obtain good empirical performance. We adapt the well-known deterministic planning system FF (Hoffmann & Nebel, 2001) to perform a suitable variant of AO* (Nilsson, 1969, 1971). We run its heuristic function on non-deterministic actions simply by acting as if we could choose the outcome, i.e., by applying the "all-outcomes determinization" (Yoon, Fern, & Givan, 2007). We run large-scale experiments with this modified FF, on the full SAM model as used in SAP. We show that runtime performance is satisfactory in the vast majority of cases; we point out the remaining challenges.

We have also integrated our planning technology into two BPM process modeling environments, making the planning functionality conveniently accessible for non-IT users. Processes (and plans) in these environments are displayed in a human-readable format. Users can specify status variable values, for example the planning goal, in simple intuitive drop-down menus. One of the environments is integrated as a research extension into the commercial SAP NetWeaver platform. Having said that, our technology is not yet part of an actual SAP product; we will discuss this in Section 7.

The treatment of non-deterministic actions, in our formalism and algorithms, is specific to our application context. This notwithstanding, it is plausible that these techniques could be useful also in other applications dealing with such actions. From a more general perspective, the contribution of our work is (A) pointing out that it is possible to leverage software-engineering models for planning, and (B) demonstrating that such an application can be realized at one of the major players in the BPM industry, thus providing a large-scale case study. The principle underlying SAM – modeling software artifacts at a level of abstraction corresponding to business users – is not limited to SAP. Thus our work may inspire similar approaches in related contexts.

We next give a brief background on planning and BPM. We then discuss the SAM model in Section 3, explaining its structure, its context at SAP, and the added value of using it for planning. We design our planning formalization in Section 4, explain our planning algorithms in Section 5, and evaluate these experimentally in Section 6. Section 7 describes our prototypes at SAP. Section 8 discusses related work, and Section 9 concludes.

## 2. Background

We introduce the basic concepts relevant to our work. We start with planning, then overview business process management (BPM) and its connection to planning.

### 2.1 Planning

There are many variants of planning (for an overview, see Traverso, Ghallab, & Nau, 2005). To handle SAM, we build on a wide-spread classical planning framework, planning with finite-domain variables (e.g., Bäckström & Nebel, 1995; Helmert, 2006, 2009). We will extend that framework with a particular kind of "non-deterministic" actions, whose semantics relates to notions from planning under uncertainty that we outline below.

**Definition 1 (Planning Task)** *A* finite-domain planning task *is a tuple* $(X, A, I, G)$. $X$ *is a set of* variables; *each* $x \in X$ *is associated with a finite domain* $dom(x)$. $A$ *is a set of* actions, *where each* $a \in A$ *takes the form* $(pre_a, eff_a)$ *with* $pre_a$ *(the* precondition*) and* $eff_a$ *(the* effect*) each being a partial variable assignment.* $I$ *is a variable assignment representing the* initial state, *and* $G$ *is a partial variable assignment representing the* goal.

A *fact* is a statement $x = c$ where $x \in X$ and $c \in dom(x)$. We identify partial variable assignments with conjunctions (sets, sometimes) of facts in the obvious way. A *state* $s$ is a complete variable assignment. An action $a$ is *applicable* in $s$ iff $s \models pre_a$. If $f$ is a partial variable assignment, then $s \oplus f$ is the variable assignment that coincides with $f$ on each variable where $f$ is defined, and that coincides with $s$ on the variables where $f$ is undefined.

**Definition 2 (Plan)** *Let $(X, A, I, G)$ be a finite-domain planning task. Let $s$ be a state, and let $T$ be a sequence of actions from $A$. We say that $T$ is a* solution *for $s$ iff either:*

*(i) $T$ is empty and $s \models G$; or*

*(ii) $T = \langle a \rangle \circ T'$, $s \models pre_a$, and $T'$ is a solution for $s \oplus eff_a$.*

*If $T$ is a solution for $I$, then $T$ is called a* plan.

One can, of course, define plans for finite-domain planning tasks in a simpler way; the present formulation makes it easier to extend the definition later on. We remark that, despite the simplicity of this formalism, it is **PSPACE**-complete to decide whether or not a plan exists (this follows directly from the results in Bylander, 1994).

Unlike in classical planning, there exist *disjunctive effects* in SAM, i.e., actions that have more than one possible outcome. This type of situation is dealt with in *planning under uncertainty*. To model SAM's disjunctive effects appropriately, we will need a mixture of what is known as *non-deterministic actions* (e.g., Smith & Weld, 1999) and what is known as *observation actions* (e.g., Weld, Anderson, & Smith, 1998).

Non-deterministic actions $a$ are like usual actions except that, in place of a single effect $eff_a$, they have a set $E_a$ of such effects, referred to as their possible *outcomes*. Whenever we apply $a$ at plan execution time, any one of the outcomes in $E_a$ will occur; separate applications of $a$ are independent. For example, $a$ might throw a dice. At plan generation time, we do not know which outcome will occur, so we must "cater for all cases". The most straightforward framework for doing so is *conformant planning* (e.g., Smith & Weld, 1999), where the plan is still a sequence of actions, and is required to achieve the goal no matter what outcomes occur during execution. Note that this does not exploit observability, i.e., the plan does not make case distinctions based on what outcomes actually do occur. To handle SAM, we will include such case distinctions, along the lines of what is known as *contingent planning* (e.g., Weld et al., 1998). In that framework, case distinctions are made by explicit observation actions in the plan. Typically, an observation action $a$ observes the – previously unknown – value of a particular state variable $x$ at plan execution time (for example, the value of a dice after throwing it). The plan branches on all possible values of $x$, i.e., $a$ has one successor for each value in $dom(x)$. Thus the plan is now a tree of actions, and the requirement is that the goal is fulfilled in every leaf of that tree.

The most wide-spread input language for planning systems today is the *Planning Domain Definition Language (PDDL)*, as used in the international planning competitions (IPC) (McDermott et al., 1998; Bacchus, 2000; Fox & Long, 2003; Hoffmann & Edelkamp, 2005; Younes, Littman, Weissman, & Asmuth, 2005; Gerevini, Haslum, Long, Saetti, & Dimopoulos, 2009). We do not get into the details of this language, since for our purposes here PDDL is merely a particular syntax for implementing our formalisms. More important for us, regarding the usability of our PDDL encoding of SAM, is the fact that PDDL has a lot of variants, with varying degrees of support by existing planning systems. Our PDDL syntax for SAM is in the PDDL variant used in the non-deterministic tracks of the IPC, i.e., the tracks dealing with non-probabilistic planning under uncertainty (Bonet & Givan, 2006; Bryce & Buffet, 2008). Specifically, we use only the most basic PDDL constructs (often referred to as "STRIPS"), except in action preconditions where we use quantifier-free formulas (Pednault, 1989; Bacchus, 2000). This PDDL subset is supported by most existing planners, in particular all those based on FF (Hoffmann & Nebel, 2001) or Fast Downward (Helmert, 2006). The limiting factor for planner support are the non-deterministic actions, for which we use the most common syntax, namely the "(oneof $eff_1 \ldots eff_n$)" construct from the non-deterministic IPC. Non-deterministic actions are supported by only few planners. Further, the semantics we will give to plans using these actions, as fits our application based on SAM, is non-standard and not supported by existing planners. This notwithstanding, several existing approaches are closely related (cf. Section 8), and, as we show herein, at least one planner – Contingent-FF (Hoffmann & Brafman, 2005) – can be adapted quite easily and successfully to deal with the new semantics.

## 2.2 Business Process Management

According to the commonly used definition (e.g., Weske, 2007), *a business process consists of a set of activities that are performed in coordination in an organizational and technical environment. These activities jointly realize a business goal.* In other words, business processes are how enterprises do business. *Business process models* serve as an abstraction of the way enterprises do business. For example, a business process model may specify which steps are taken, by various entities across an enterprise, to send out a customer quote answering a request for quotation. The atomic steps in such a process model may be both, manual steps performed by employees, or automatic steps executed on the IT infrastructure. We will refer to process models simply as "processes".

An explicit model of processes allows all sorts of support and automation, addressed in the area of *business process management (BPM)*. Herein, we are mostly concerned with process creation and adaptation. That is done in *BPM modeling environments*. Importantly, the users of these environments will typically not be IT experts, but *business experts* – the people familiar with, and taking decisions for, the business. The dominant paradigm for representing business processes are *workflows*, also called *control-flows*, often formalized as Petri nets (e.g., Aalst, 1997). Such a control-flow defines an order of execution for the process steps, within certain degrees of flexibility implied, for example, by parallelism. For business experts, the control-flow is displayed in a human-readable format, typically a flow diagram. Our application at SAP uses Business Process Modeling Notation (Object Management Group, 2008), short *BPMN*, which we will illustrate in Section 7.

An alternative paradigm for representing business processes, which relates to the SAM model we consider herein, are *constraint-based representations* (e.g., Wainer & de Lima Bezerra, 2003; van der Aalst & Pesic, 2006; Pesic, Schonenberg, Sidorova, & van der Aalst, 2007). These model processes implicitly through their desired properties, rather than explicitly through concrete workflows. This kind of representation is more flexible, in that, by modifying the model, we can modify the entire process space. For example, we might add a new constraint "archive customer quotes only if all follow-ups have been created". Such a representation is also more explicit about the reasons for process design, supporting human understanding. The downside is that, for actual automated process execution, a concrete control-flow design is required. One way of viewing our planning technology is that it provides the service of generating such control-flow designs for SAM.

Processes are executed on IT infrastructures, like the one provided by SAP. Such execution coordinates the individual processing steps, prompting human users as appropriate, and performing all the necessary data updating on IT level. This is realized in dedicated *process execution engines* (Dumas, ter Hofstede, & van der Aalst, 2005). Clearly, the execution poses high demands on the structure of the workflow. The most basic requirement is that the atomic process steps correspond to actual steps known to the IT infrastructure.[3]

The requirements on business processes, such as legal and financial regulations, are subject to frequent updates. The people responsible for adapting the processes – business experts – are not familiar with the IT infrastructure, and may come up with processes whose "atomic steps" are nowhere near what can be implemented easily, partially overlap with whole sets of existing functions, and/or require the implementation of new functions although existing functions could have been arranged to do the job. Thus there is a need for intensive communication between business experts and IT experts, incurring significant costs for human labor and increased time-to-market.

How can planning come to the rescue? As indicated, the basic (and well-known) idea is to use a planning tool for composing (an approximation of) the process automatically, helping the business expert to come up with a process close to the IT infrastructure. The main novelty in our work is that we leverage a pre-existing model, SAM, getting us around one of the most critical issues in the area: the overhead for creating the planner input.

## 3. SAM

We explain the structure of the SAM language, and give a running example. We outline the background of SAM at SAP, and explain the added value of using SAM for planning.

### 3.1 SAM Structure and Example

Status and Action Management (SAM) models belong to business objects (BOs). Each BO is associated with a set of finite-domain "status" variables, and with a set of actions. Each status variable highlights one value that the variable will take when a new instance of the BO is created. Each action is described with a textual label (its name), a precondition, and an effect. The precondition and effect are propositional formulas over the variable values.

**Definition 3 (SAM BO)** *A SAM business object $o$ is a triple $(X(o), A(o), I(o))$. $X(o)$ is a set of* status variables*; each $x \in X(o)$ is associated with a finite domain $dom(x)$. $A(o)$ is a set of* actions*, where each $a(o) \in A(o)$ takes the form $(pre_{a(o)}, eff_{a(o)})$; $pre_{a(o)}$ (the* pre*-condition) is a propositional formula over the atoms $\{x = c \mid x \in X(o), c \in dom(x)\}$; $eff_{a(o)}$ (the* effect*) is a negation-free propositional formula over these same atoms, in disjunctive normal form (DNF). $I(o)$ is a variable assignment representing $o$'s initial state.*

This structure is in obvious correspondence with that of Definition 1. The only differences are that there is no "goal", and that the preconditions and effects are more complex. In our planning application, the goal is set by the user creating a new process. We discuss in Section 4 how to extend Definitions 1 and 2 to handle SAM preconditions and effects.

Note that there are no cross-BO constraints in SAM – each BO $o$ refers only to values of its own variables. This is a shortcoming of the current version of SAM: in reality, BOs do interact. We will get back to this further below.

---

3. Another important requirement is an appropriate *data-flow* (van der Aalst, 2003; Dumas et al., 2005), e.g., sending a manager the documents required to decide whether or not to accept a customer quote. Since, in our case, the data is encapsulated into business objects, this is not a major issue for us.

| Action name | precondition | effect |
|---|---|---|
| Check CQ Completeness | CQ.archiving:notArchived | CQ.completeness:complete OR CQ.completeness:notComplete |
| Check CQ Consistency | CQ.archiving:notArchived | CQ.consistency:consistent OR CQ.consistency:notConsistent |
| Check CQ Approval Status | CQ.archiving:notArchived AND CQ.approval:notChecked AND CQ.completeness:complete AND CQ.consistency:consistent | CQ.approval:necessary OR CQ.approval:notNecessary |
| Decide CQ Approval | CQ.archiving:notArchived AND CQ.approval:necessary | CQ.approval:granted OR CQ.approval:notGranted |
| Submit CQ | CQ.archiving:notArchived AND (CQ.approval:notNecessary OR CQ.approval:granted) | CQ.submission:submitted |
| Mark CQ as Accepted | CQ.archiving:notArchived AND CQ.submission:submitted | CQ.acceptance:accepted |
| Create Follow-Up for CQ | CQ.archiving:notArchived AND CQ.acceptance:accepted | CQ.followUp:documentCreated |
| Archive CQ | CQ.archiving:notArchived | CQ.archiving:archived |

Figure 1: Our SAM-like running example, modeling the behavior of "customer quotes" CQ.

For illustration, Figure 1 gives a SAM-like model for a BO called "customer quote (CQ)", that will be our running example. For confidentiality reasons, the shown object and model are artificial, i.e., they are *not* contained in SAM as used at SAP. By "CQ.x:c" we denote the proposition $x = c$, in the object CQ. The initial state $I(CQ)$ is:

- "CQ.archiving:notArchived",

- "CQ.completeness:notComplete",

- "CQ.consistency:notConsistent",

- "CQ.approval:notChecked",

- "CQ.submission:notSubmitted",

- "CQ.acceptance:notAccepted",

- "CQ.followUp:documentNotCreated".

When using this example below, where relevant we will assume that the goal entered by the user is "CQ.followUp:documentCreated AND CQ.archiving:archived".

The reader should keep in mind that this is merely an illustrative example, which necessarily is simple. In particular, the intended life-cycle workflow is rather obvious, given the action descriptions in Figure 1. This is very much not the case in general. The Business Objects modeled in SAM have up to 15 status variables, yielding up to 12 million possible states (combinations of variable values) even for a single BO. In other words, SAM is a flexible model – after all, that was its main design purpose – and describes a large number of combination possibilities in a compact way. Furthermore, in two of the application scenarios for planning ((A) and (C) in Section 3.3 below), we are actually looking not for entire life-cycles but for process fragments that may begin or end at any BO status values.

## 3.2 SAM@SAP

SAM was created by SAP as part of the development of the IT infrastructure supporting SAP Business ByDesign. That infrastructure constitutes a fully-fledged SAP application. Its key advantage over traditional SAP applications is a higher degree of flexibility, facilitating the use of SAP software as-a-service. Individual system functions are encapsulated as software services, using the service-oriented architectures paradigm (Krafzig et al., 2005; Bell, 2008). The software services may be accessed from standard architectures like BPM process execution engines, thus enabling their flexible combination with other services. To further support flexibility, the Business ByDesign IT infrastructure is model-driven. IT artifacts at various system levels are described declaratively using SAP-proprietary modeling formats. Business objects are one such IT artifact, and SAM is one such format.

The original purpose of SAM was to facilitate the design, and the management of changes, during the development of the Business ByDesign infrastructure (a formidably huge enterprise). Of course, SAM also serves the implementation of changes to the infrastructure later on, should changes be required. New developments are first implemented and tested on the model level. Then parts of the program code are automatically generated from the model. Straightforward code skeletons contain the status variables, as well as function headers for the available actions (similar to what Eclipse does for Java class definitions). In addition, the skeletons are filled with code fragments performing the precondition checks and updates on status variables. Changes pertaining to the status variable level can thus be implemented in SAM models, and automatically propagated into the code. In this sense, the original semantics of SAM is as follows:

(I) When a BO $o$ is newly created, the values of the status variables are set to $I(o)$.

(II) BO actions $a(o)$ whose precondition $pre_{a(o)}$ is not fulfilled are either disallowed, or raise an exception when executed; which one is true depends on the part of the architecture attempting to execute the action.

(III) Upon execution of an action $a$, the status variables change their values as prescribed by one of the disjuncts in the effect DNF $eff_{a(o)}$. The only aspect controlled outside SAM is which disjunct is chosen: that choice is made based on BO data content not reflected in SAM.

The intention behind SAM is to formulate complex business-level dependencies between individual processing steps, using simple modeling constructs that facilitate easy modification. The formulation in terms of preconditions and effects relative to high-level status variable values was adopted as a natural means to meet these requirements. Of course, this design also took some inspiration from traditional software modeling paradigms (Schneider, 2001; Object Management Group, 2006).

Leveraging SAM for planning is a great opportunity because of the effort it takes to build such a model. SAM was developed continuously along with Business ByDesign, across a time span of more than 5 years. Throughout this time, around 200 people were involved (as a part-time occupation) in the development. SAP implemented a dedicated graphical user interface for this development. There are design patterns for typical cases, there are naming conventions, there is a fully-fledged governance process, and there even is educational training for the developers. A council of senior SAP architects supervises the development.

### 3.3 Applications of SAM-Based Planning

The Business ByDesign infrastructure is designed to be very general and adaptable, covering the needs of a great breadth of different SAP customers' business domains. To adapt the infrastructure to their practice, SAP customers may choose to create their own processes as compositions of the functionalities provided (as Web services), in a way tailored to their needs. Indeed, a second motivation behind SAM, beside its role for software development, was to facilitate such flexibility, by describing the possible process space in a declarative manner, rather than imposing standard workflows as is a common methodology in other contexts such as artefact-centric business process modeling (e.g., Cohn & Hull, 2009). SAM shares this motivation with constraint-based process representation languages. It also shares their downside, in that the actual workflows still need to be created. In this context, there are at least three application scenarios for planning based on SAM:

**(A) Development based on SAM.** During model-driven development based on SAM, planning enables developers to examine how their changes affect the process space. This greatly facilitates experimentation and testing. For example, planning can be used for debugging, testing whether or not the goal can still be reached, or whether the changes opened any unintended possibilities, like, reaching an undesired state of the BO (e.g., "CQ.consistency:notConsistent AND CQ.acceptance:accepted"). More than such reachability testing (essentially a model checking task), planning serves to generate entire processes, which as we shall see take the form of BPMN process models with parallelism and conditional splits. Developers can examine the space of processes generated in this way, determining for different combinations of start/end conditions how these can be connected. Note that the generality offered by the planning approach is an absolute requirement here – the process generation tool must be at least as general as SAM, handling propositional formula preconditions and effects.

**(B) Designing extended/customized processes.** Individual SAP customers have individual requirements on their processes, and thus may use the same BOs in different ways. For example, even if the end state of customer quotes (which in practice are much more complex than our illustrative example) always involved being archived, different businesses may differ on the side conditions: one organization only archives POs if all follow-ups have been created; another archives only POs that were successful; a third organization archives POs immediately and automatically after getting a response; a fourth only based on an explicit user-request. Part of the motivation behind SAM is to provide such flexibility. Planning based on SAM can be used to automatically generate a first version of the desired process.[4]

**(C) Process redesign.** Sometimes the best option is to design a new process from scratch. If the business experts doing so are not aware the underlying IT infrastructure, then this incurs huge costs at process implementation time. SAM opens the possibility for business experts to "explain" the individual steps in the new process in terms of status variable value changes, i.e., in terms of a start/end state corresponding to what the business user considers to be an atomic processing step. Planning then shows if and how these status changes can be implemented using existing transactions. In particular, the planner can be called for some business object X (e.g., a sales order) from within a process being created for some other object Y (e.g., a customer quote). Hence, despite the mentioned absence of cross-BO constraints in the current version of SAM, planning can help to create non-trivial processes spanning several BOs.

---

4. The alternative – equipping each BO with a standard life-cycle or a set thereof – would come at the prize of a flexibility loss for complex BOs, and is not the choice made by SAP.

All these use cases are supported by our prototype at SAP; we will illustrate its use for (C), in a cross-BO situation as mentioned, in Section 7.3.

An obvious requirement for the planner to be useful is instantaneous response. Typically a user will be sitting at the computer and waiting for the planner to answer. Further, all functionality must be accessible conveniently. In particular, each time a user wants to call the planner, she needs to provide the planning goal (and possibly the initial state). It is essential that this can be done in a simple and intuitive manner, without in-depth expertise in IT or about the BO the question. Thus we limit ourselves to conjunctive goals in the sense of "I want these status variables to have these values at the end of the process", like the goal "CQ.followUp:documentCreated AND CQ.archiving:archived" in our illustrative example. In our prototype, such goals are specified using simple drop-down menus.

SAM was not originally intended to do planning, and is of course not perfect for that purpose. We will discuss the main limitations in Section 9, but we need to briefly touch on two points here already. The absence of cross-BO constraints in the current version of SAM has implications for planner setup and performance, and will play a role in our experiments.[5] Another issue is plan quality. The duration/cost of the actions may differ vastly, but SAM does not contain any information about this: it is not relevant to SAM's original purpose, software engineering. We will not address plan quality measures herein. Our planning algorithm of course attempts to find small plans. But it gives no quality guarantee in that regard, and the practical value of such a guarantee would be doubtful.

## 4. Planning Formalization

We design the syntax and semantics of a suitable planning formalism capturing SAM, and we illustrate that formalism using our running example.

### 4.1 SAM Planning Tasks: Syntax

Given the close correspondence of SAM business objects (Definition 3) with finite-domain planning tasks (Definition 1), it is straightforward to extend the latter to capture the former.

**Definition 4 (SAM Planning Task)** *A SAM planning task is a tuple $(X, A, I, G)$ whose elements are the same as in finite-domain planning tasks, except for the action set $A$. Each $a \in A$ takes the form $(pre_a, E_a)$ with $pre_a$ being a propositional formula over the atoms $\{x = c \mid x \in X, c \in dom(x)\}$, and $E_a$ being a set of partial variable assignments. The members $\mathit{eff}_a \in E_a$ are the outcomes of $a$.*

As discussed above, we keep the goal as simple as possible. For the effects, in place of the negation-free propositional DNF formulas of Definition 3, we now have sets $E_a$ of outcomes. The action preconditions are as in Definition 3. This generalizes the partial variable assignments from Definition 1 – which are equivalent to negation-free conjunctions over the atoms $\{x = c \mid x \in X, c \in dom(x)\}$ – to arbitrary propositional formulas over these atoms. That generalization poses no issue for defining the plan semantics; at implementation level, most current planning systems compile such preconditions into negation-free conjunctions, using the methods originally proposed by Gazen and Knoblock (1997).

---

5. As we shall discuss in Section 9, BO interactions do exist. An according extension of SAM is planned, which could in principle be tackled using the exact same planning technology as presented herein.

To obtain a SAM planning task $(X, A, I, G)$, when given as input a SAM business object $o = (X(o), A(o), I(o))$ along with a goal conjunction $G(o)$, we first set $X := X(o)$, $I := I(o)$, and $G := G(o)$. For each $a(o) \in A(o)$ we include one $a$ into $A$, where $pre_a := pre_{a(o)}$. As for $\mathit{eff}_{a(o)}$, we create one partial variable assignment $\mathit{eff}_a$ for each disjunct in that DNF formula, and we define the possible outcomes $E_a$ as the set of all these $\mathit{eff}_a$.

By convention, we denote with $A^d := \{a \in A \mid |E_a| = 1\}$ and $A^{nd} := \{a \in A \mid |E_a| > 1\}$ the sets of *deterministic* and *non-deterministic* actions of a SAM planning task, respectively. If $a \in A^d$, then by $\mathit{eff}_a$ we denote the single outcome of $a$.

## 4.2 SAM Planning Tasks: Semantics

SAM action preconditions $pre_{a(o)}$ are in direct correspondence with usual planning preconditions, cf. point (II) in Section 3.2. By contrast, SAM's disjunctive effects $\mathit{eff}_{a(o)}$ require to create a mix of two different kinds of planning actions – non-deterministic actions and observation actions – from the literature. To understand this, reconsider the role of SAM action effects $\mathit{eff}_{a(o)}$ in their original environment, i.e., point (III) in Section 3.2. Any one of the disjuncts will occur, and at plan generation time we do not know which one. At plan execution time, the SAP system executing the action will observe the relevant data content, and will decide which branch to take. In the example from Figure 1, "Check CQ Completeness" will answer "CQ.completeness:complete" if the BO data is complete, and will answer "CQ.completeness:notComplete" otherwise. Of course, the SAP system keeps track of which outcomes occured. In other words, (a) SAM's disjunctive effects correspond to observation actions, that (b) internally observe environment data not modeled at the planning level. Due to (a), it makes perfect sense to handle such actions by introducing case distinctions at plan generation time, one for each outcome. Due to (b), there is no direct link of the "observation" to a reduction of uncertainty at planning level. During execution, the values of the "observed variables" are known prior to the "observation" already, and change as a result of applying that action. For example, "CQ.completeness.notComplete" is considered to be true prior to the first application of "Check CQ Completeness", and may be changed to "CQ.completeness.complete" by that action. In that respect, and in that the outcome set (an arbitrary DNF) is more general than the domain of a particular variable, SAM's disjunctive effects are more similar to the common notions of non-deterministic actions.

For simplicity, we will henceforth refer to SAM's disjunctive-effects actions as non-deterministic actions. Another important point regarding these actions is that data content is not allowed to change while the process is running; the data is filled in directly upon creation of the BO. Thus the outcome of a non-deterministic action will be the same throughout the plan execution, and it makes no sense to execute such an action more than once in a plan. For example, there is no point in repeatedly applying "Check CQ Completeness".

A final issue is to decide what a "plan" actually is. Cimatti, Pistore, Roveri, and Traverso (2003) describe the three most common concepts, in the presence of non-deterministic ac-tions: *strong plans*, *strong cyclic plans*, and *weak plans*. We will discuss the latter two below; the most desirable property is the first one. A strong plan guarantees to reach the goal no matter which action outcomes occur. We now define this formally, for our setting.

An *action tree over $A$* is a tree whose nodes are actions from $A$, and whose edges are labeled with partial variable assignments. Each action $a$ in the tree has exactly $|E_a|$ outgoing edges, one for (and labeled with) each $\mathit{eff}_a \in E_a$. In the following definitions, $A_{av}^{nd}$ refers to the subset of non-deterministic actions that have not yet been used, and are thus still available, at any given state during plan execution. Recall that $s \oplus \mathit{eff}_a$, defined in Section 2, over-writes $s$ with those variable values defined in $\mathit{eff}_a$, and leaves $s$ unchanged elsewhere.

**Definition 5 (Strong SAM Plan)** *Let $(X, A, I, G)$ be a SAM planning task with $A = A^d \cup A^{nd}$. Let $s$ be a state, let $A_{av}^{nd} \subseteq A^{nd}$, and let $T$ be an action tree over $A \cup \{STOP\}$. We say that $T$ is a* strong SAM solution *for $(s, A_{av}^{nd})$ iff either:*

*(i) $T$ consists of the single node STOP, and $s \models G$; or*

*(ii) the root of $T$ is a $\in A^d$, $s \models \mathit{pre}_a$, and the sub-tree of $T$ rooted at $a$'s child is a strong SAM solution for $(s \oplus \mathit{eff}_a, A_{av}^{nd})$; or*

*(iii) the root of $T$ is a $\in A_{av}^{nd}$, $s \models \mathit{pre}_a$, and, for each of $a$'s children reached via an edge labeled with $\mathit{eff}_a \in E_a$, the sub-tree of $T$ rooted at that child is a strong SAM solution for $(s \oplus \mathit{eff}_a, A_{av}^{nd} \setminus \{a\})$.*

*If $T$ is a strong solution for $(I, A^{nd})$, then $T$ is called a* strong SAM plan.

Compare this to Definition 2. Item (i) of the present definition is essentially the same, saying that there is nothing to do if the goal is already true. In difference to Definition 2, we then distinguish deterministic actions (ii) and non-deterministic ones (iii). In the former case, $a$ has a single child and we require the remainder of the tree to solve that child, similarly as in Definition 2. In the latter case, $a$ has several children all of which need to be solved by the respective sub-tree. This corresponds to the desired case distinction observing action outcomes at plan execution time.

Note that, throughout the plan, there is no uncertainty about the current variable values. Note also that we solve, not a state, but a pair consisting of a state and a subset of non-deterministic actions. This reflects the fact that whether or not an action tree solves a state depends not only on the state itself, but also on which non-deterministic actions are still available. The maintenance of the set $A_{av}^{nd}$ ensures that we allow each non-deterministic action only once, on each path through $T$ (but the action may occur several times on separate paths). Thus any one execution of the plan applies the action at most once.

The problem with Definition 5 is that strong plans typically do not exist. To illustrate this, consider Figure 2, showing a *weak SAM plan*, a notion we will now formally define, for our running example from Figure 1. Recall that the goal is assumed to be "CQ.followUp:documentCreated AND CQ.archiving:archived". If either of "Check CQ Completeness" or ""Check CQ Consistency", as shown at the top of Figure 2, result in a negative outcome ("CQ.completeness:notComplete" or "CQ.completeness:notConsistent"), then the goal becomes unreachable. Thus a strong plan does not exist for this SAM planning task. That phenomenon is not limited to this illustrative example. In our experiments, almost 75% of a very large sample of SAM planning tasks did not have a strong plan.

To address this, one can define more complicated goals, or a weaker notion of plans. For the former option, one could use goals specifying alternatives, preferences, and/or temporal plan properties (e.g., Pistore & Traverso, 2001; Dal Lago, Pistore, & Traverso, 2002; Shaparau, Pistore, & Traverso, 2006; Gerevini et al., 2009). However, goals will be specified online by business users and it is absolutely essential for this to be as simple as possible. We hence decided to go for the second option.[6]

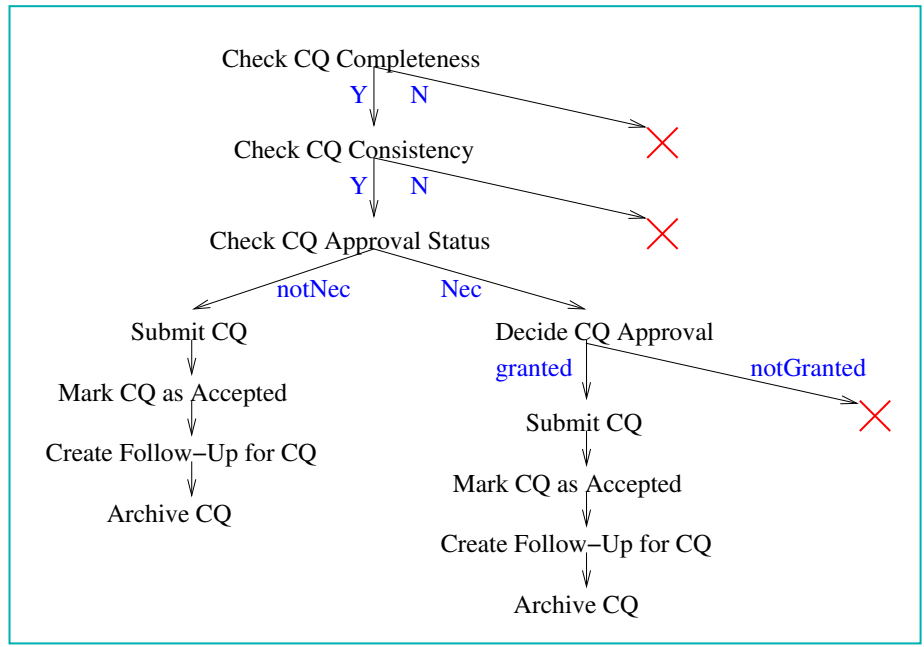

Figure 2: A weak SAM plan for the running example from Figure 1. *STOP* actions not shown, *FAIL* actions marked by (red) crosses.

The weak plans of Cimatti et al. (2003) are too liberal for our purposes. They guarantee only that at least one possible execution of the plan reaches the goal, posing no requirements on all the other executions. For example, in Figure 2, this would mean to allow the plan to handle only the left-hand side outcome of "Check CQ Approval Status", i.e., "CQ.approval:notNecessary", and to do nothing at all about (attach the empty tree at) its other outcome, "CQ.approval:necessary".

So what about strong cyclic plans? There, the plan may have cycles, provided every plan state *can*, in principle, reach the goal. This allows to "wait for a desired outcome", like a cycle around a dice throw, waiting to obtain a "6". Alas, repetitions of non-deterministic SAM actions will always produce the same outcome. It is futile to insert a cycle at the top of Figure 2, waiting for the desired outcome of "Check CQ Completeness". While it is plausible to prompt a user to edit the BO content and then repeat the check (placeholders for such cycles could be inserted as a planning post-process), this is not a suitable exception handling in general. Exception handling depends on the business context, and typically depends on the actual customer using the SAP system. This is impossible to reflect in a model maintained centrally by SAP.

In conclusion, from the perspective of SAM-based planning there is not much one can do other than to highlight the bad outcomes to the user, so that the exception handling can

---

6. Some works on more complex goals can be employed to define alternative notions of "weak plans" (by using trivial fall-back goals). We will discuss this in some detail in Section 8.

be inserted manually afterwards. Of course, a non-deterministic action should have at least one successful outcome, or else it would be completely displaced in a process. Further, it is essential to highlight outcomes as "bad" only if they really are "bad", i.e., to not mark as failed any outcomes that could actually be solved. Our definition reflects all this:

**Definition 6 (Weak SAM Plan)** *Let $(X, A, I, G)$ be a SAM planning task with $A = A^d \cup A^{nd}$. Let $s$ be a state, let $A_{av}^{nd} \subseteq A^{nd}$, and let $T$ be an action tree over $A \cup \{STOP, FAIL\}$. We say that $T$ is a weak SAM solution for $(s, A_{av}^{nd})$ iff either:*

  *(i) $T$ consists of the single node STOP, and $s \models G$; or*

  *(ii) the root of $T$ is a $a \in A^d$, $s \models pre_a$, and the sub-tree of $T$ rooted at $a$'s child is a weak SAM solution for $(s \oplus eff_a, A_{av}^{nd})$; or*

  *(iii) the root of $T$ is a $a \in A_{av}^{nd}$, $s \models pre_a$, and, for each of $a$'s children reached via an edge labeled with $eff_a \in E_a$, we have that either: (a) the sub-tree of $T$ rooted at that child is a weak SAM solution for $(s \oplus eff_a, A_{av}^{nd} \setminus \{a\})$; or (b) the sub-tree of $T$ rooted at that child consists of the single node FAIL, and there exists no action tree $T'$ that is a weak SAM solution for $(s \oplus eff_a, A_{av}^{nd} \setminus \{a\})$; where (a) is the case for at least one of $a$'s children.*

*If $T$ is a weak solution for $(I, A^{nd})$, then $T$ is called a weak SAM plan.*[7]

Compared to Definition 5, the only difference lies in item (iii), which no longer requires every child to be solved. Instead, the arrangement of options (a) and (b) means that failed nodes – leaves in the tree that stop the plan without success – are tolerated, as long as at least one child is solved, and every failed node is actually unsolvable. This is in obvious correspondence with our discussion above. In Figure 2, the failed nodes, i.e., the sub-trees consisting only of the special *FAIL* action, are crossed out (in red). Note the difference to Cimatti et al.'s (2003) definition of "weak plans" discussed above: we are not allowed to cross out the right-hand side outcome of "Check CQ Approval Status", i.e.,"CQ.approval:necessary", because that outcome is solvable.

We remark that allowing non-deterministic actions only once (or, more generally, having an upper bound on repetition of non-deterministic actions) is required for Definition 6 to make sense. In item (iii), the definition recurses on itself when stating that some children may be unsolvable. While such recursion occurs also at other points in Definitions 5 and 6, at those points the action tree $T$ considered is reduced by at least one node. For unsolvable children of non-deterministic actions in Definition 6 (iii) (b), such a reduction is not given – the quantification is over *any* action tree $T'$ that may be suitable to solve the child. What makes the recursion sound, instead, is that the set of available non-deterministic actions is diminished by one. Without this, the notion of "weak SAM plan" would be ill-defined: the recursion step may result in the same planning task over again, allowing the construction of planning tasks that are considered "solvable" if they are "unsolvable".[7]

---

7. Concretely, say we obtain Definition 6' from Definition 6 by considering states $s$ only, removing the handling of $A_{av}^{nd}$. Consider the example with one variable $x$ whose possible values are $A$ and $B$, with initial state $I : x = A$, with goal $G : x = B$, and with a single action $a$ with two possible outcomes, $x = A$ or $x = B$. Say $T$ consists only of $a$. Then the "bad" outcome of $a$, i.e., the state $x = A$, is identical to the original initial state $I$. If $I$ is "unsolvable" according to Definition 6', then this outcome of $a$ qualifies for Definition 6' (iii) (b), and thus the overall task – the same state $I$ – is considered to be "solvable". By contrast, using Definition 6 as above, the plan must solve the state/available-non-deterministic-actions pair $(x = A, \{a\})$, and the bad outcome of $a$ is the *different* pair $(x = A, \emptyset)$. That pair is unsolvable, and hence $a$ is a weak plan.

The following observation holds simply because Definition 5 captures a special case of Definition 6:

**Proposition 1 (Weak SAM Plans Generalize Strong SAM Plans)** *Let $(X, A, I, G)$ be a SAM planning task with $A = A^d \cup A^{nd}$. Let $s$ be a state, let $A_{av}^{nd} \subseteq A^{nd}$, and let $T$ be an action tree over $A \cup \{STOP\}$. If $T$ is a strong SAM solution for $(s, A_{av}^{nd})$, then $T$ is a weak SAM solution for $(s, A_{av}^{nd})$.*

In other words, any strong SAM plan is also a weak SAM plan, and hence in particular any SAM planning task that is solvable under the strong semantics is also solvable under the weak semantics. The inverse is obviously not true. A counter-example is our running example in Figure 2.

We remark that, trivially, deciding whether a plan exists is hard for both, Definition 5 and Definition 6. The special case where all actions are deterministic is a generalization of Definition 2, where as mentioned that problem is **PSPACE**-complete.

## 4.3 SAM Planning Tasks: Running Example

For illustration, we encode our running example, Figure 1, into a SAM planning task $(X, A, I, G)$. We set $X := \{Arch, Compl, Cons, Appr, Subm, Acc, FoUp\}$, abbreviating the status variable names mentioned in Figure 1. For example, $Arch$ stands for the variable "CQ.archiving". The domain of each of $Arch$, $Compl$, $Cons$, $Subm$, $Acc$, and $FoUp$ is $\{true, false\}$. This serves to abbreviate the various names used for the respective variable values in Figure 1. The domain of $Appr$ is $\{notChecked, nec, notNec, granted, notGranted\}$. In what follows, for brevity we write facts, i.e., variable/value pairs, involving true/false valued variables like literals. For example, we write $\neg FoUp$ instead of $(FoUp, no)$.

The initial state of the SAM BO, and thus the SAM planning task, is:

- $I = \{\neg Arch, \neg Compl, \neg Cons, (Appr, notChecked), \neg Subm, \neg Acc, \neg FoUp\}$

The goal is "CQ.followUp:documentCreated AND CQ.archiving:archived":

- $G = \{FoUp, Arch\}$

The deterministic actions $A^d$ are:

- "Mark CQ as Accepted": $(\neg Arch \wedge Subm, \{Acc\})$
- "Create Follow-Up for CQ": $(\neg Arch \wedge Acc, \{FoUp\})$
- "Archive CQ": $(\neg Arch, \{Arch\})$
- "Submit CQ": $(\neg Arch \wedge ((Appr, notNec) \vee (Appr, granted)), \{Subm\})$

Note here that action effects are sets of partial variable assignments, i.e., sets of sets of facts. For the deterministic actions, there is just one partial variable assignment so we omit the second pair of set parentheses to avoid notational clutter. Note also that we do not have "delete effects". The effects assign new values to the affected variables, implicitly removing their old values, cf. the meaning of $s \oplus eff_a$ as defined in Section 2.

The non-deterministic actions $A^{nd}$ are:

- Check CQ Completeness: $(\neg Arch, \{\{Compl\}, \{\neg Compl\}\})$
- Check CQ Consistency: $(\neg Arch, \{\{Cons\}, \{\neg Cons\}\})$

- Check CQ Approval Status:

  $(\neg Arch \wedge (Appr, notChecked) \wedge Compl \wedge Cons, \{\{(Appr, nec)\}, \{(Appr, notNec)\}\})$

- Decide CQ Approval:

  $(\neg Arch \wedge (Appr, nec), \{\{(Appr, granted)\}, \{(Appr, notGranted)\}\})$

Figure 2 shows a weak SAM plan for this example. For presentation to the user, a simple post-process (outlined in Section 7.1) transforms such plans into BPMN workflows.

## 5. Planning Algorithms

We design an adaptation of FF (Hoffmann & Nebel, 2001), using a variant of the AO* forward search from Contingent-FF (Hoffmann & Brafman, 2005), as well as a naïve extension of FF's heuristic function. We assume that the reader is familiar with heuristic search in general, and we refer to the literature (e.g., Pearl, 1984) for that background.

### 5.1 Search

For strong SAM planning – Definition 5 – we use AO* tree search (Nilsson, 1969, 1971). For weak SAM planning – Definition 6 – we use a variant of that search that we refer to as *SAM-AO\**. We focus in what follows mainly on SAM-AO*, since AO* is well-known and will become clear as a side effect of the discussion.

Search is forward in an AND-OR tree whose nodes are states (OR nodes) and actions (AND nodes). The OR'ed children of states are the applicable actions, the AND'ed children of actions are the alternative outcomes (for deterministic actions, there is a single child so the AND node trivializes). Like in AO*, we propagate "node solved" and "node failed" markers. The mechanics for this are the usual ones in the case of OR nodes, i.e., the marker of the node is the disjunction of its childrens' markers. For AND nodes, SAM-AO* differs from the usual conjunctive interpretation, implementing the weak SAM planning semantics of Definition 6: amongst other things, an AND node is failed only if *all* its children are failed. Figure 3 provides an overview of SAM-AO*, highlighting the differences to AO*. Figure 4 illustrates the algorithm on a simplification of our running example.

One feature of the algorithm that is immediately apparent is the book-keeping which non-deterministic actions are still available. Recall here that, in line with Definitions 5 and 6, we allow each non-deterministic action at most once in any execution of a plan. For the search algorithm – both in strong planning (AO*) and in weak planning (SAM-AO*) – this means that OR nodes contain not only a state $s$, but a pair $(s, A_{av}^{nd})$ giving the state as well as the subset of $A^{nd}$ that has not been used up to this node. We will refer to such pairs as *search states* from now on. The book-keeping of the sets $A_{av}^{nd}$ is straightforward. For the initial state, all non-deterministic actions are still available. Whenever a non-deterministic action $a$ is applied, for its outcome states, $a$ is no longer available. For illustration, consider how the action sets are reduced in Figure 4 (B–D).

The heuristic function $h$, that we assume as a given here, takes as arguments the search state, i.e., both the state and the available non-deterministic actions. This is because action availability affects goal distance and hence the heuristic estimates. By $h(s) = 0$ the heuristic indicates goal states, and by $h(s) = \infty$ it may indicate that the state is unsolvable. The algorithm trusts the heuristic, i.e., it assumes that $h$ only returns these values if the state

**procedure** *SAM-AO\**

**input** SAM planning task $(X, A, I, G)$ with $A = A^d \cup A^{nd}$, heuristic function $h$

**output** A weak plan for $(X, A, I, G)$, or "unsolvable"

initialize $T$ to consist only of $N_I$; $content(N_I) := (I, A^{nd})$

$status(N_I) := solved$ if $h(I, A^{nd}) = 0$, $failed$ if $h(I, A^{nd}) = \infty$, $unknown$ else

**while** $status(N_I) = unknown$ **do**

  $N_s := select\text{-}open\text{-}node(T)$; $(s, A_{av}^{nd}) := content(N_s)$

  **for** all $a \in A^d \cup A_{av}^{nd}$ with $s \models pre_a$ **do**

    **if** $a \in A^d$ and *is-direct-duplicate*$(N_s, s \oplus \mathit{eff}_a, A_{av}^{nd})$ **then** skip $a$ **endif**

    insert $N_a$ as child of $N_s$ into $T$; $content(N_a) := a$

    $A_{av}^{nd\prime} := A_{av}^{nd}$ if $a \in A^d$, else $A_{av}^{nd\prime} := A_{av}^{nd} \setminus \{a\}$

    **for** all $\mathit{eff}_a \in E_a$ **do**

      $s' := s \oplus \mathit{eff}_a$

      insert $N_{s'}$ as child of $N_a$ into $T$; $content(N_{s'}) := (s', A_{av}^{nd\prime})$

      $status(N_{s'}) := solved$ if $h(s', A_{av}^{nd\prime}) = 0$, $failed$ if $h(s', A_{av}^{nd\prime}) = \infty$, $unknown$ else

    **endfor**

    $status(N_a) := SAM\text{-}aggregate(\{status(N') \mid N' \text{ is child of } N_a \text{ in } T\})$

  **endfor**

  $status(N_s) := OR\text{-}aggregate(\{status(N') \mid N' \text{ is child of } N_s \text{ in } T\})$

  *propagate-status-updates-to-I*$(N_s)$

**endwhile**

**if** $status(N_I) = failed$ **then return** "unsolvable" **endif**

**return** an action tree corresponding to a subtree $T'$ of $T$ s.t. $N_I \in T'$ and:

      for all inner nodes $N_s \in T'$: $status(N_s) = solved$ and $N_s$ has exactly one child $N_a$ in $T'$;

      for all nodes $N_a \in T'$: all children $N_{s'}$ of $N_a$ in $T$ are contained in $T'$

$is\text{-}direct\text{-}duplicate(N, s', A_{av}^{nd}) := \begin{cases} true & \exists \text{ predecessor } N_0 \text{ of } N \text{ s.t. } content(N_0) = (s', A_{av}^{nd}) \\ false & \text{else} \end{cases}$

$SAM\text{-}aggregate(M) := \begin{cases} solved & \exists m \in M : m = solved, \text{ and} \\ & \forall m \in M : (m = solved \text{ or } m = failed) \\ failed & \forall m \in M : m = failed \\ unknown & \text{else} \end{cases}$

$OR\text{-}aggregate(M) := \begin{cases} solved & \exists m \in M : m = solved \\ failed & \forall m \in M : m = failed \\ unknown & \text{else} \end{cases}$

Figure 3: Pseudo-code of SAM-AO\*, highlighting the differences to AO\*.

$s$ is indeed a goal state/unsolvable. Detecting unsolvable states is within the capabilities of FF's heuristic, and is of paramount importance for planning with SAM models. Its behavior is like that of the heuristic in Figure 4 (B–D), which immediately marks all unsolvable nodes as being such. We come back to this in Section 5.2.2 below.

The overall structure of SAM-AO\* is the same as that of AO\*. Starting with the initial search state (compare Figure 4 (A)), we iteratively use *select-open-node* to select an OR node $N_s$ in the tree that has not yet been expanded and whose status is unknown; the selection criterion is based on the node's *f-value*, as explained below. We expand the selected node with the applicable actions (in Figure 4 (B), we omit "Check CQ Consistency" to save space), and we insert one new node for each possible outcome of these actions ("Comp" vs.

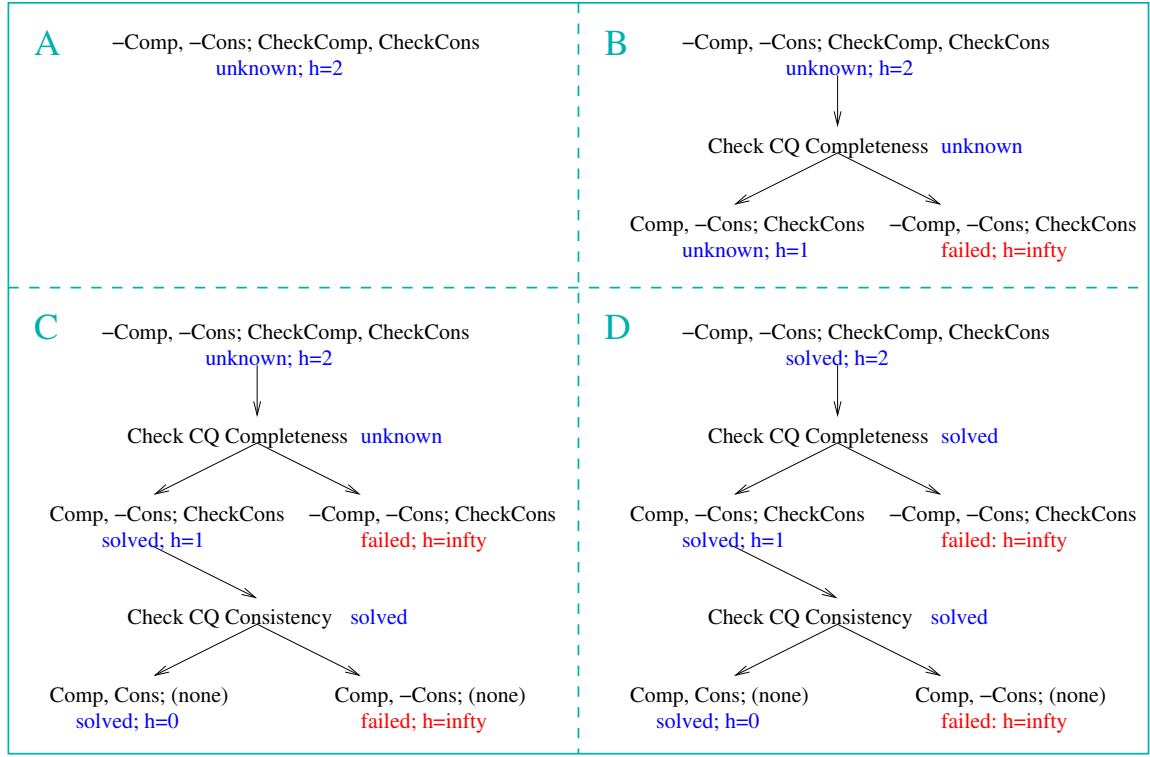

Figure 4: Phases of SAM-AO* in a simplifaction of the running example (Figure 1), with only the two variables "CQ.completeness" and "CQ.consistency", and only the two actions "Check CQ Completeness" and "Check CQ Consistency". The goal, using the abbrevations here, is "Comp, Cons". For search states $(s, A_{av}^{nd})$, on the left-hand side of the semicolon we show the state $s$, and on the right-hand side we show the set $A_{av}^{nd}$ of available non-deterministic actions.

"-Comp" in Figure 4 (B)); we will discuss the *is-direct-duplicate* function below. Each new node, i.e., the corresponding search state, is evaluated with the heuristic function. Once all outcomes of an action $a$ were inserted, the status of $a$ is updated. Once all actions applicable to the current node $N_s$ are inserted, the status of $N_s$ is updated. The latter update, reflected in the *OR-aggregate* equation in Figure 3, is exactly as in AO*. A key difference to AO* lies in the former update, reflected in the *SAM-aggregate* equation. That equation is in obvious correspondence with Definition 6. In Figure 4 (B), neither of these updates yields any new information, because the status of one of the action outcomes, "Comp, Cons; CheckCons", is "*unknown*". That changes in Figure 4 (C), where the status of both outcomes becomes definite (solved/failed), and the updates propagate this information to the action $a$ and the search state node $N_s$ it was applied to.

After the status of $N_s$ and $a$ has been set, *propagate-status-updates-to-I*$(N_s)$ performs a backward iteration starting at $N_s$, updating each action and search state along the way to $I$ using the same two functions, *OR-aggregate* and *SAM-aggregate*. This is necessary since the status of $N_s$ may have changed, and that may affect the status of any predecessors. This happens, for example, in Figure 4 (D) where the status of the first node and action now change from "*unknown*" to "*solved*". The algorithm terminates when the initial node (the search tree root) is solved or failed. In the former case, a solved sub-tree is returned. That happens in Figure 4 (D), and the sub-tree returned is equivalent to the start (top two actions) of our example plan in Figure 2.

In addition to status markers, SAM-AO* also annotates search states with their $f$-values, as well as the current best action. This is not shown in Figure 3 since it is (almost) identical to what is done in AO*. The $f$-value of a search state node is the minimum of those of its children, plus 1 accounting for the cost of applying an action; a minimizing child is the best action. The $f$-value of an action node is the maximum of its children, except – and herein lies the only difference to AO* – that we do not set the action value to $\infty$ unless all its children are marked as failed. The *select-open-node* procedure starts in $N_I$ and keeps choosing best actions until it arrives at a non-expanded state, which is selected as $N_s$.

The *is-direct-duplicate* function in Figure 3 disallows the generation of search states that are identical to one of their predecessors in the tree. We refer to this as *direct duplicate pruning*. Note that the method prunes duplicates only within deterministic parts of the search tree. If a predecessor node $N_0$ as in Figure 3 is found, then all actions between $N_0$ and $N$ are deterministic, because otherwise $content(N_0)$ would contain strictly more non-deterministic actions than $N$. Obviously, direct duplicate pruning preserves soundness and completeness of the search algorithm. We have:

**Proposition 2 (SAM-AO* is Complete and Sound)** *Let $(X, A, I, G)$ be a SAM planning task, and let $h$ be a heuristic function. SAM-AO* terminates when run on the task with $h$. Provided $h(s) = 0$ iff $s$ is a goal state, and $h(s) = \infty$ only if $s$ is unsolvable, SAM-AO* terminates with success iff there exists a weak SAM plan for the task, and the action tree returned in that case is such a plan.*

This follows from the known results about AO*, by definition, and by two simple observations. First, eventually, on any tree path no non-deterministic actions will be available anymore. Second, direct duplicate pruning allows only finitely many nodes in a SAM planning task without non-deterministic actions.

The reader might wonder whether stronger duplicate pruning methods could be defined, across the non-deterministic actions in the tree. A naïve approach, asking only whether a predecessor of $N$ contains the same state – and ignoring the sets of available non-deterministic actions – does not work for SAM-AO*. It renders that algorithm unsound. This is because such a pruning method may mark solvable search states as failed, and failed nodes can be part of the solution in a weak plan. For illustration, consider the simple example where we have a variable $x$ with values $A, B, C$, the initial state $A$, the goal $C$, and three actions: $a_1$ has precondition $A$ and the two possible outcomes $B$ and $C$; $a_2$ has precondition $B$ and outcome $A$; $a_3$ has precondition $A$ and outcome $C$. Say the search has chosen to apply $a_1$ first. Consider $a_1$'s unfavorable outcome $B$. At this point, in order to obtain a plan, we must apply $a_2, a_3$ to achieve the goal $C$. However, the outcome state $s : x = A$ of $a_2$ is the same as the initial state. Hence $s$ is pruned, hence $a_1$'s outcome $B$ is marked as failed, hence the algorithm wrongly concludes that $a_1$'s outcomes qualify for Definition 6 (iii), and that $a_1$ on its own is a plan.

## 5.2 Heuristic Function

To compute goal distance estimates, we use "all-outcomes-determinization" as known in probabilistic planning (Yoon et al., 2007) to get rid of non-deterministic actions, then run the FF heuristic (Hoffmann & Nebel, 2001) off-the-shelf. For the sake of self-containedness, we next explain this in some detail. The reader familiar with FF may skip to Section 5.2.2.

### 5.2.1 RELAXED PLANNING GRAPHS

FF's heuristic function is one out of a range of general-purpose planning heuristics based on a relaxation widely known as "ignoring delete lists" (McDermott, 1999; Bonet & Geffner, 2001). Heuristics of this kind have emerged in the late 90s and are still highly successful. In what follows, we assume that action preconditions and the goal are conjunctions of positive atoms, and are thus equivalent to sets of facts. For more general formulas, such as the preconditions in SAM planning tasks, one can apply known transformations (Gazen & Knoblock, 1997) to achieve this.

The name "delete lists" comes from a Boolean-variable representation of planning tasks. Translated to our context, the relaxation means that variables accumulate, rather than change, their values. For illustration, say we have "CQ.archiving:notArchived" in our running example, and we apply the action "Archive CQ" whose effect is "CQ.archiving:archived". Then, in the relaxation, the resulting state will be "CQ.archiving:notArchived, CQ.archiving:archived", containing *both* the old and the new value of the variable "CQ.archiving". Thus the other actions of this BO, that all require the customer quote to not be archived yet, remain applicable, and in difference to the plan from Figure 2, a relaxed plan can archive the CQ right at the start and then proceed to the rest of the processing.

Viewing variable assignments as sets of facts, relaxed action application is equivalent to taking the set union of the current state with the action effect. This yields a strictly larger set of facts than the real application of the action ("CQ.archiving:notArchived, CQ.archiving:archived" instead of "CQ.archiving:archived"). Satisfaction of preconditions and the goal is then tested by asking for inclusion in that set. Bylander (1994) proved that, within the relaxation, plan existence can be decided in polynomial time. He also proved, however, that optimal relaxed planning, i.e., finding the length of a shortest possible relaxed plan, is still **NP**-hard. Therefore, the heuristics used by practical planners approximate that length. Specifically, the FF heuristic we build on herein computes some not necessarily optimal relaxed plan. The algorithm doing so consists of two phases. First, it builds a *relaxed planning graph (RPG)* to approximate forward reachability. Then it extracts a relaxed plan from the RPG.

Figure 5 shows how an RPG is computed in our planner. The algorithm gets the state $s$ as well as the remaining non-deterministic actions, $A_{av}^{nd}$. It then determinizes the latter actions, inserting each of their possible outcomes as an individual new deterministic action into the new action set $A'$. The following loop is a simple fixed point operation over sets of facts. The initial set $F_0$ is equal to the state $s$ whose goal distance shall be estimated. Each loop iteration then increments $F_t$ with the effects of all actions whose preconditions have been reached. In case all goals are reached, the algorithm stops with success and returns the iteration index $t$. If a fixed point occurs before that happens, the algorithm returns $\infty$.

```
procedure RPG
    input SAM planning task (X, A, I, G) with A = A^d ∪ A^{nd},
          state s, available non-deterministic actions A^{nd}_{av}
    output Number of relaxed parallel steps needed to reach the goal, or ∞
    A' := A^d ∪ {(pre_a, {eff_a}) | a ∈ A^{nd}_{av}, eff_a ∈ E_a}
    F_0 := s, t := 0
    while G ⊄ F_t do
        A'_t := {a ∈ A' | pre_a ⊆ F_t}
        F_{t+1} := F_t ∪ ⋃_{a∈A'_t} eff_a
        if F_{t+1} = F_t then return ∞ endif
        t := t + 1
    endwhile
    return t
```

Figure 5: Pseudo-code for building a relaxed planning graph (RPG).

If the RPG returns $t < \infty$, the heuristic function algorithm enters its second phase, relaxed plan extraction. This is a straightforward backchaining procedure selecting supporting actions for the goals, and then iteratively for the supporting actions' preconditions. The backchaining makes sure to select only feasible supporters by exploiting the reachability information encoded in the sets $F_t$. Note here that $t$ itself is not a good heuristic estimator because it counts *parallel* action applications – we could make transactions on 1000 BOs in parallel and still count this as a single step.

Consider our simplified example from Figure 4. For the root node "Comp, Cons; Check-Comp, CheckCons", the relaxed plan returned will be ⟨"Check CQ Completeness$^+$", "Check CQ Consistency$^+$"⟩, where the superscript "+" indicates that these are actions from the determinized set $A'$ in Figure 5, choosing the positive outcome of each of these actions. The heuristic value returned is 2, as in Figure 4. Indeed, all the heuristic values from Figure 4 are as would be returned by FF's heuristic. In particular, if "-Comp", i.e., "CQ.completeness:notComplete", holds in a state, but the action "Check CQ Completeness" is no longer available, then the heuristic value returned is $\infty$ because no action in $A'$ can achieve the goal "Comp", and thus the RPG fixed point does not contain the goal. Similarly if "-Cons" holds in a state but "Check CQ Consistency" is no longer available.

### 5.2.2 Detecting Failed Nodes

It follows directly from, e.g., the results of Hoffmann and Nebel (2001), that the RPG stops with success iff there exists a relaxed plan for the task $(X, A', s, G)$. From this, we easily get the following result which is relevant for us:

**Proposition 3 (RPG Dead-End Detection in SAM is Sound)** *Let $(X, A, I, G)$ be a SAM planning task with $A = A^d \cup A^{nd}$. Let $s$ be a state, and let $A^{nd}_{av} \subseteq A^{nd}$ be a set of non-deterministic actions. If the RPG run on these inputs returns $\infty$, then there exists no weak SAM solution for $(s, A^{nd}_{av})$.*

To see this, note that the action set $A'$ of Figure 5 is, from the perspective of plan existence, an over-approximation of the actual action set $A^d \cup A^{nd}_{av}$ we got available. $A'$ allows us to choose, for any non-deterministic action, the outcome that we want. Thus, from a plan using $A^d \cup A^{nd}_{av}$ we can trivially construct a plan using $A'$. So if no plan using $A'$ exists then neither does a plan using $A^d \cup A^{nd}_{av}$. From here it suffices to see that non-existence of a relaxed plan (based on $A'$) implies non-existence of a real plan (based on $A'$). That is obvious, concluding the argument.

It is of course a very strong simplification to act as if one could choose the outcomes of non-deterministic actions. Part of our motivation for doing so is to demonstrate that it is not necessary, at least in this application context, to dramatically enhance off-the-shelf planning techniques. The simplistic approach just presented suffices to obtain good performance. This is particularly true regarding the ability to detect dead-ends. We experimented with a total of 548987 planning instances based on SAM. Of these (within limited time/memory) we found a weak plan for 441884 instances. Around half of the actions in these plans are non-deterministic, and these typically yield failed nodes in the plan. For every one of these failed nodes, in every one of the 441884 solved instances, the RPG returned $\infty$.

### 5.2.3 Helpful Actions Pruning

We also adopt FF's *helpful actions pruning*. Aside from the goal distance estimate, the relaxed plan can be used to determine a most promising subset $H(s)$ – the helpful actions – of the actions applicable to the evaluated state $s$. Essentially, $H(s)$ consists of the actions that are applicable to $s$ and that are contained in the relaxed plan computed as described above in Section 5.2.1.[8] This action subset is used as a pruning method simply by restricting, during search, the expansion of state $s$ to consider only the actions $H(s)$. This kind of heuristic action pruning is of paramount importance for planner performance (Hoffmann & Nebel, 2001; Richter & Helmert, 2009).

In the SAM setting, one important aspect of FF's helpful actions pruning is that it is accurate enough to distinguish relevant BOs from irrelevant ones. That is to say, if a BO is not mentioned in the goal, then no action pertaining to it will ever be considered to be helpful. This is simply because, as pointed out previously, SAM currently does not model cross-BO interactions. So if a BO $Y$ is not in the goal then relaxed plan extraction will never create any sub-goals pertaining to $Y$.

The obvious – and well-known – caveat of helpful actions pruning is that it does not preserve completeness. $H(s)$ may not contain any of the actions that actually start a plan from $s$. If that happens, then search may stop unsuccessfully even though a plan exists. This pertains to classical planning just as it pertains to AO* and SAM-AO* as used herein.

Importantly, helpful actions pruning in SAM-AO*, i.e., for weak SAM planning as per Definition 6, has another more subtle caveat: *it does not preserve soundness*. Consider again the example where we have a variable $x$ with values $A, B, C$, the initial state is $A$, the goal is $C$, and we have three actions of which action $a_1$ has precondition $A$ and two possible outcomes $B$ and $C$, $a_2$ has precondition $B$ and outcome $A$, and $a_3$ has precondition $A$ and outcome $C$. Say the search has applied $a_1$. Say that $N_s := (s, A^{nd}_{av})$ is the node corresponding to $a_1$'s unfavorable outcome $B$. The only way to complete $a_1$ into a plan is to attach $a_2, a_3$ to $N_s$. Presume that helpful actions pruning, at the node $N_s$, removes $a_2$. Then $N_s$ is marked as failed, and we wrongly conclude that $a_1$ on its own is a plan.

---

8. FF's definition of $H(s)$ is a little more complicated, adding also some actions that were not selected for the relaxed plan but achieve a relevant sub-goal. We omit this for brevity. Recent variants of helpful actions pruning, for different heuristic functions like the causal graph heuristic (Helmert, 2006), do not make such additions, selecting $H(s)$ based on membership in abstract solutions only.

Using helpful actions pruning, one may incorrectly mark a node $N_s$ as failed. If such $N_s$ is a leaf in a weak plan $T$, marked as failed even though it is solvable by action tree $T'$, then $T$ is not a valid plan. This can be fixed, in a plan-correction post-process, by attaching $T'$ to $N_s$, where $T'$ is found by running SAM-AO* without helpful actions pruning on $N_s$. We did not implement such a post-process because, according to our experiments, it is unnecessary in practice: as discussed at the end of the previous sub-section, all failed nodes $N_s$ in our 441884 weak plans have heuristic value $\infty$, and are thus proved to be, indeed, unsolvable.

# 6. Experiments

We will describe our prototype at SAP in the next section. In what follows, we evaluate our planning techniques in detail from a scientific point of view. Our experiments are aimed at understanding three issues:

(1) *What is the applicability of strong respectively weak planning in SAM?*

(2) *Is the runtime performance of our planner sufficient for the envisioned application?*

(3) *How interesting is SAM as a planning benchmark?*

We first explain the experiments setup. We then describe our experiments with FF for strong plans, and with FF for weak plans; we summarize our findings with blind search. While these experiments consider instances pertaining to a single BO, we finally examine what happens when scaling the number of relevant BOs.

## 6.1 Experiments Setup

All experiments were run on a 1.8 GHz CPU, with a 10 minute time and 0.5 GB memory cut-off. Our planner is implemented in C as a modification of FF-v2.3. The source code, the problem generator used in our experiments, and the anonymized PDDL encoding of SAM are available for download at `http://www.loria.fr/~hoffmanj/SAP-PDDL.zip`. Our SAM-AO* implementation is modifed from the AO* implementation in Contingent-FF (Hoffmann & Brafman, 2005). Like that planner, we weight heuristic values by a factor of 5 (we did not play with this parameter).

We focus on the case where the initial state is set as specified in SAM. Thus a SAM planning instance in what follows is identified by its goal: a subset of variable values. The number of such instances is finite, but enormous; just for choosing the subset of variables to be constrained we have $2^{1110}$ options. In what follows, we mostly consider goals all of whose variables belong to a single BO. This is sensible because, as previously stated, SAM currently does not reflect interactions across BOs. We made an instance generator that allows to create instance subsets characterized by the number $|G|$ of variables constrained in the goal (this parameter is relevant for business users, and as we shall see it also heavily influences planner performance). For given $|G|$, the generator enumerates all possible variable tuples, and allows to randomly sample for each of them a given number $S$ of value tuples. The maximum number of variables of any BO, in the current version of SAM, is 15. We created all possible instances for $|G| = 1, 2, 3, 13, 14, 15$ where the number of instances is up to around 50000. For all other values of $|G|$, we chose a value for $S$ so that we got around 50000 instances each. The total number of instances we generated is 548987.

Since SAM currently does not model cross-BO interactions, for a single-BO goal we can in principle supply to the planner only those actions pertaining to that BO. We will henceforth refer to this option as using the *BO-relevant* actions. Contrasting with this, the *full* actions option supplies to the planner all actions (no matter which BO they pertain to). We will use the BO-relevant actions in some experiments where we wish to enable the planner to prove the planning task to be unsolvable – with the full actions, this is always impossible because the reachable state space is much too vast. In our baseline, however, we use the full actions. The motivation for this is that helpful actions pruning will detect the irrelevant actions anyway (cf. Section 5.2), and in the long term, it is likely that SAM will model cross-BO interactions.

## 6.2 Strong SAM Plans

In our first experiment, we evaluate the performance of strong planning on SAM, i.e, we run FF in a standard AO* tree search forcing all children of AND nodes to be solved. We identify two parameters relevant to the performance of FF: the kind of BO considered, and $|G|$. Figure 6 shows how coverage and state evaluations (number of calls to the heuristic function) depend on these parameters.

Consider first Figure 6 (a). The $x$-axis ranges over BOs, i.e., each data point corresponds to one kind of BO.[9] The ordering of BOs is by decreasing percentage of solved instances. For each BO, the $y$-axis shows the percentage of solved, unsolved, and failed instances within that BO. The overall message is mixed. On the one hand, coverage is perfect for 194 of the 371 BOs, so for more than half of the BOs all the tested instances have a strong plan which is found by FF. On the other hand, for the other 177 BOs, coverage is rather bad. For 51 of the BOs, not a single instance is solved. On the 126 BOs in between, coverage declines steeply. Importantly, when counting unsolved cases in total (across BOs), it turns out that 88.83% of the instances are unsolved. In other words, **for almost 90% of the tested cases FF's search space does not contain a strong SAM plan**. Of course, this percentage pertains to the particular distribution of test cases that we used. Still this result indicates that the applicability of strong planning in SAM is quite limited.

Another interesting aspect of Figure 6 (a) is that failed cases are rare: they constitute only 0.8% of the total instance set. That is, due to helpful actions pruning, FF's search spaces are typically small enough to be exhausted within the given time and memory.

Consider now Figure 6 (b), which shows coverage on the $y$-axis over $|G|$ on the $x$-axis. Again, the message is mixed. On the one hand, with a single goal ($|G| = 1$), 58.85% of the instances are solved with a strong plan. On the other hand, the number of solved cases declines monotonically, and quite steeply, over growing $|G|$. With 2 goals we are at 36.95%, with 4 goals at 29.03%, with 5 goals at 23.86%. For $|G| \geq 10$, the number of solved cases is less than 5%, and for $|G| \geq 13$, the number is less than 1%.

One may wonder at this point whether it is FF's helpful actions pruning that is responsible for the frequent non-existence of strong plans. The answer is "no". In a second experiment with strong planning, we ran FF without helpful actions, giving as input only the BO-relevant actions in order to enable proofs of unsolvability. The result is very clear: the number of solved cases hardly changes at all. The total percentage of solved cases in the previous experiment is 10.38%, the total percentage in the new experiment is 10.36%. This low success rate is due to unsolvability, not to prohibitively large search spaces. In total, 74.1% of the instances are proved unsolvable; FF fails only on 15.54%.

---

9. Note that we do not include all 404 BOs. Precisely, we consider 371 of them. The remaining 33 BOs are not interesting for planning: all variable values not true in the initial state are unreachable because these values are set by procedures not encoded in SAM.

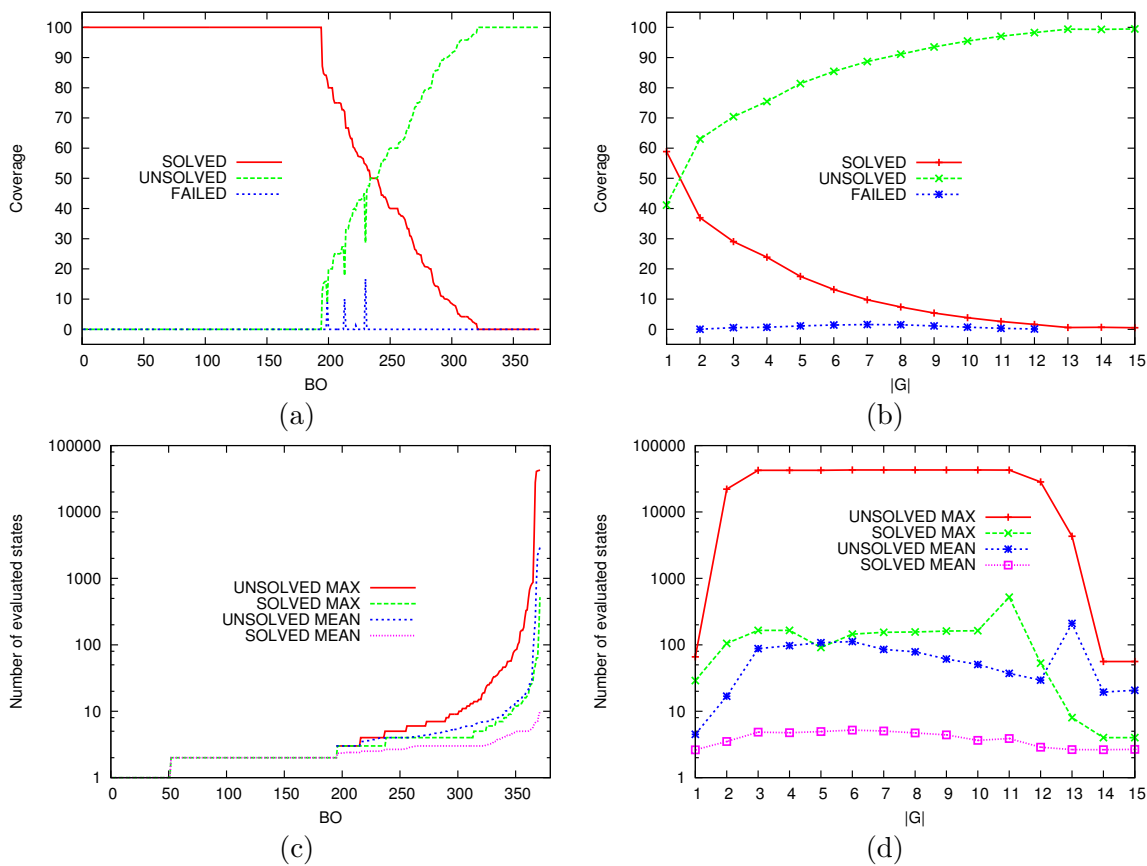

Figure 6: **Strong planning with FF on full action sets**. Coverage (a,b) and state evaluations (c,d) data, plotted over individual kinds of BOs (a,c) and $|G|$ (b,d). "SOLVED": plan found. "UNSOLVED": search space (with helpful actions pruning) exhausted. "FAILED": out of time or memory. Ordering of BOs in (c) is by increasing $y$-value for each curve individually.

Given the above, the applicability of strong planning in SAM appears limited unless we can restrict attention to BOs with few variables and/or to single-goal planning tasks. From a more general perspective, the best option appears to be to:

(I) *Try to find a strong SAM plan (using FF with AO\*).*

(II) *If (I) fails, try to find a weak SAM plan (using FF with SAM-AO\*).*

In this setting, it is relevant how long we will have to wait for the answer to (I). Figures 6 (c) and (d) provide data for this. We consider the instances where FF terminated regularly (plan found or helpful actions search space exhausted), and we consider performance in terms of the number of evaluated states, i.e., the number of calls to the heuristic function.

The ordering of BOs in Figure 6 (c) is by increasing $y$-value for each curve individually; otherwise the plot would be unreadable. The most striking observation is that, for 351 of the 371 BOs, the maximum number of state evaluations is below 100. For solved instances, this even holds for 369 BOs, i.e., for all but 2 of the BOs. The maximum number of state evaluations done in order to find a plan is 521, taking 0.22 seconds total runtime. The mean behavior is even more good-natured, peaking at 9.85 state evaluations. Waiting for a "no" can be more time consuming, with a peak at 42954 evaluations respectively 110.87 seconds. However, since all the "yes" answers are given very quickly, for practical use in an online business process modeling environment it seems feasible to simply give the strong planning one second (or less), and switch to weak planning in case that was not successful.

Consider Figure 6 (d). Like in (c), we can observe the very low number of state evaluations required for the solved cases. Somewhat surprisingly, there is no conclusive behavior over $|G|$. The reasons are not entirely clear to us. The "UNSOLVED MAX" curve is flat at the top because larger search spaces lead to failure. The discontinuities around $|G| = 12$ are presumably due to BO structure. Few BOs have more than 12 variables, so the variance in the data is higher in this region. For the sharp drops in "UNSOLVED MAX" and "SOLVED MAX", an additional factor is that there are very few strong plans for such large goals (cf. Figure 6 (b)): those strong plans that do exist are found easily; disproving existence of a strong plan can be easier for larger goals, since this increases the chance that the relaxed plan will identify at least one unsolvable goal.

Summing up our findings regarding the issue (1) [applicability of strong vs. weak planning in SAM] we wish to understand in these experiments, SAM does not admit many strong plans, but those instances that do have them tend to be solved easily by FF.

## 6.3 Weak SAM Plans

We will now see that **weak SAM planning can solve 8 times as many instances as strong SAM planning** – namely around 80% of our test cases. Precisely, of the 548987 instances, 441884 are solved; all but 43 of these are solved by the default configuration of our planner. The average percentage of non-deterministic actions, across all weak plans, is 48.29%; the maximum percentage is 91.67%. Figure 7 shows our results, giving the same four kinds of plots as previously shown for strong planning in Figure 6.

Consider first Figure 7 (a). We see that, now, coverage is perfect in 274 of the 371 kinds of BOs, as opposed to the 194 BOs of which that is true for strong planning. The latter BOs are a subset of the former: wherever strong planning has perfect coverage the same is true of weak planning. Whereas strong planning has 0 coverage – no instance solved at all – for 51 BOs, we have no such cases here. The minimum coverage is 18.07%, and coverage is below 50% only for 9 BOs. In total, while strong planning solves only 10.38% of the test cases, we now solve 80.48%. That said, we still have 17.12% unsolved cases and 2.4% failed cases, and this gets much worse for some BOs. Per individual BO, the fraction of unsolved instances peaks at 81.92%, and the fraction of failed instances peaks at 14.98%.

Consider now Figure 7 (b). $|G| = 1$ is handled perfectly – 100% coverage as opposed to 58.85% for strong planning – but this is followed by a fairly steady decline as $|G|$ grows. An explanation for the discontinuity at $|G| = 3, 4$ could be that for $|G| = 3$ our experiment is exhaustive while for $|G| = 4$ we only sample. As above, the higher variance for large $|G|$ can be explained by the much smaller number of BOs in this region.

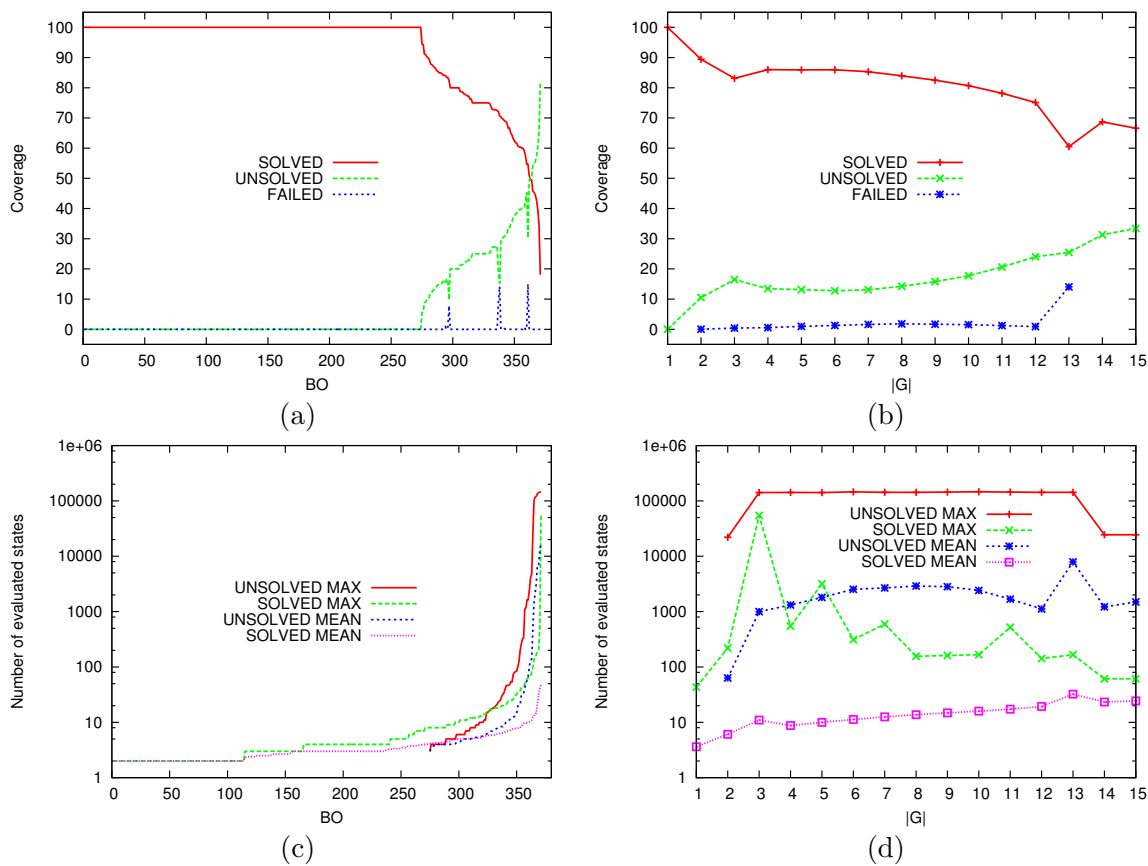

Figure 7: **Weak planning with FF on full action sets**. Coverage (a,b) and state evaluations (c,d) data, plotted over individual kinds of BOs (a,c) and $|G|$ (b,d). "SOLVED": plan found. "UNSOLVED": search space (with helpful actions pruning) exhausted. "FAILED": out of time or memory. Ordering of BOs in (c) is by increasing $y$-value for each curve individually.

Figures 7 (c) and (d) provide a deeper look into performance on those instances where FF terminated regularly (plan found or helpful actions search space exhausted). Like in Figure 6 (c), the ordering of BOs in Figure 7 (c) is by increasing $y$-value for each curve individually. The number of state evaluations is typically low. The phenomenon is not quite as extreme as shown for strong planning in Figure 6 (c). This is likely because the more generous definition of plans makes it more difficult for FF to prove AND nodes unsolvable, and hence to prune large parts of the search space. In detail, for 350 of the 371 BOs, the maximum number of state evaluations is below 100; for solved instances, this holds for 364 BOs (the corresponding numbers for strong planning are 351 and 369). The maximum number of state evaluations done in order to find a plan is 54386, taking 27.41 seconds total runtime (521 and 0.22 for strong planning). From the "SOLVED MEAN" curve we see that this maximal case is very exceptional – the mean number of state evaluations per BO peaks at 47.78. Comparing this to the "UNSOLVED MEAN" curve, we see that a large number of search nodes is, as one would expect, much more typical for unsolved instances.

In Figure 7 (d), we see that the overall behavior of state evaluations over $|G|$ largely mirrors that of coverage, including the discontinuity at $|G| = 3, 4$. The most notable exception is the fairly consistent decline of "SOLVED MAX" for $|G| > 3$. It is unclear to us what the reason for that is.

What is the conclusion regarding the issues (2) [planner performance] and (3) [benchmark challenge] we wish to understand? For issue (2), our results look fairly positive. In particular, consider only the solved instances (weak plan found). As explained above, the number of state evaluations is largely well-behaved. In addition, the heuristic function is quite fast. As stated, the maximum runtime is 27.41 seconds. The second largest runtime is 2.6 seconds, and the third largest runtime is 1.69 seconds; all other plans are found in less than 0.3 seconds. So a practical approach could be to simply apply a small cut off, like 0.5 seconds, or perhaps a minute if time is not as critical. This yields a quick step (II) as a follow-up of the similarly quick step (I) determined above for strong planning.

What this strategy leaves us with are, in total, 17.12% unsolved instances and 2.4% failed ones. Are those an important benchmark challenge for future research? Answering this question first of all entails finding out whether we can solve these instances when not using helpful actions pruning, and if not, whether they are solvable at all.

We ran FF without helpful actions pruning on the unsolved and failed instances of Figure 7, slightly more than 100000 instances in total. We enabled unsolvability proofs by giving as input only the BO-relevant actions, and we facilitated larger search spaces by increasing the time/memory cut-offs from 10 minutes and 0.5 GB to 30 minutes and 1.5 GB respectively. All failed instances are still failed in this new configuration. Of the previously unsolved instances, 47.43% are failed, and 52.52% are proved unsolvable.[10] Only 0.05% – 43 instances – are now solved (the largest plan contains 140 actions). The influence of $|G|$ and the kind of BO is similar to what we have seen. The number of state evaluations is vastly higher than before, with a mean and max of 10996.72 respectively 289484 for *solved* instances. But the heuristic is extremely fast with only the BO-relevant actions, and hence finding a plan takes a mean runtime of only 0.12 seconds. The max, second max, and third max runtimes are 2.94 seconds, 0.7 seconds, and 0.53 seconds respectively; all other plans are found in less than 0.15 seconds. Thus, with the above, all but 6 of the 441884 weak plans in this experiment are found in less than 0.5 seconds.

All in all, changing the planner configuration achieves some progress on the instances not solved by the default configuration, and it appears that many of them are unsolvable anyway. But certainly they are a challenge for further research.

## 6.4 Blind Search

To explore to what extent heuristic techniques are actually required to successfully deal with this domain – and thus to what extent the domain constitutes an interesting benchmark for such techniques – we ran an experiment with blind search. We used AO* with a trivial heuristic that returns 1 on non-goal states and 0 on goal states. Since weak planning is much more applicable than strong planning in SAM, we used the weak planning semantics. We provided as input only the BO-relevant actions – otherwise, blind forward search is trivially hopeless due to the enormous branching factor.

---

10. Unsolvability of certain goal value combinations, i.e., partial assignments to a BO's variables, occurs naturally since these variables are not independent. For example, some of the unsolvable instances required a BO to simultaneously satisfy "BO.approval:In Approval" and "BO.release:Released". In our running example, this kind of situation arises, e.g., when requiring "CQ.approval:notChecked" together with "CQ.acceptance:accepted".

For the sake of conciseness, we do not discuss the results in detail here. In summary, blind search is quite hopeless. It runs out of time or memory on 79.36% of our test instances. It solves 19.04% of them – as opposed to the 80.48% solved based on FF. Interestingly, due to FF's ability to detect dead-ends via relaxed planning graphs, blind search is worse than heuristic search even at proving unsolvability: in total, this happens in 5.99% of the cases using FF, and in 1.60% of the cases using blind search.

That said, for BOs that have only few status variables and/or status values, and for goals of size 1 or 2, blind search fares well, if not as well as heuristic search. The interesting benchmarks lie outside this region – which is the case for more than 90% of our test instances.

## 6.5 Scaling Across BOs

FF does not scale gracefully to planning tasks with several BOs. We selected, for each BO, one solved instance $m(BO)$ with maximum number of state evaluations. Since that will be of interest, we include here all 404 BOs, i.e., also those 33 BOs all of whose planning goals are trivial (either unsolvable or true in the BO's initial state already); $m(BO)$ is 1 for these BOs. We generated 404 planning tasks $COM_k$, for $1 \leq k \leq 404$, combining the goals $m(BO)$ for all BOs up to number $k$, in an arbitrary ordering of the BOs. We compared the data thus obtained against data we refer to as $ACC_k$, obtained by summing up the state evaluations when running FF in turn on each of the individual goals $m(BO)$. This comparison is valid since the BOs are mutually independent, and a plan for $COM_k$ can be obtained as the union of the plans for the individual goals. Figure 8 shows the data.

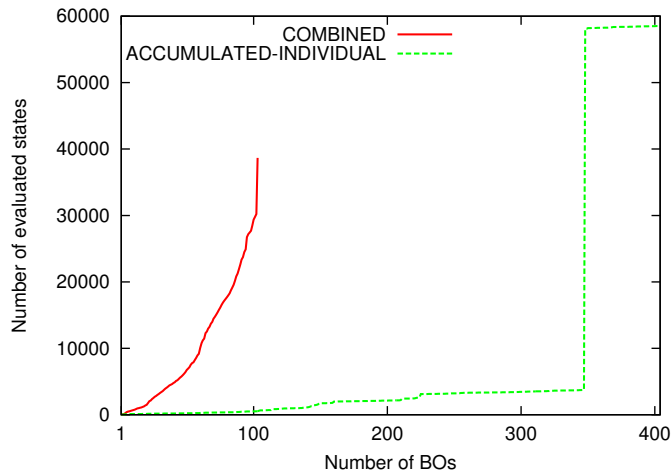

Figure 8: **Weak planning with FF when scaling the number of relevant BOs**. State evaluations plotted over the number of BOs for which a goal is specified. "COMBINED" means that FF is run on the conjunction of all these goals ($COM_k$ in the text). "ACCUMULATED-INDIVIDUAL" gives the sum of the data when running FF individually on each single goal ($ACC_k$ in the text).

As Figure 8 shows quite clearly, FF does not scale gracefully to planning tasks with several BOs.[11] The largest instance solved is $k = 103$, with 38665 evaluations. The sum of state evaluations when solving the 103 sub-tasks individually is 529. A possible explanation is that, adding more goals for additional BOs, more actions are helpful. The increased number of nodes may multiply over the search depth. Interestingly, the disproportionate search increase occurs even when the new goal added is trivial. For example, $ACC_{98}$ has just 1 more state evaluation than $ACC_{97}$, while for $COM_{98}$ and $COM_{97}$ that difference amounts to 251 state evaluations. On the other hand, for up to $k \leq 14$ BOs, $ACC_k$ is still below 1000; the difficulties arise only when $k$ becomes quite large.

## 7. Application Prototypes at SAP

We have integrated our technology into two BPM modeling environments. We next briefly explain how we transform the planner output into a BPM format. We then outline the positioning of our prototypes at SAP, and illustrate the business user view of our technology. We close with a few words on how the prototypes have been evaluated SAP-internally.

### 7.1 Transforming Plans to Business Processes

Business users expect to get a process model in a human-readable BPM workflow format. We use BPMN (Object Management Group, 2008). The BPMN process model corresponding to Figure 2 is depicted in Figure 9. This process model makes use of alternative ("x") and parallel ("+") execution, unifies redundant sub-trees ("Submit CQ" ... "Archive CQ"), removes failed outcomes, and highlights in red those nodes that may have such outcomes. These changes are obtained using the following simple post-process to planning.

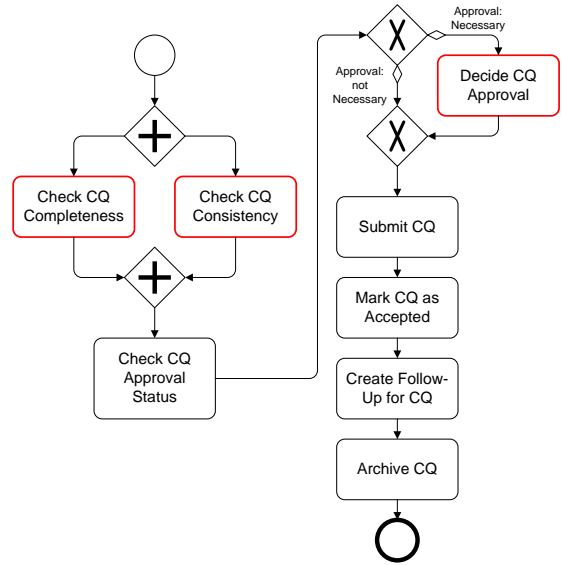

Figure 9: Final BPM process created for the running example.

---

11. The "vertical" part of the plot for $ACC_k$ is because, as we noticed before, the globally maximal number of state evaluations, 54386 for BO 347, is an extreme outlier.

First, we remove each failed node together with the edge leading to it. In our running example, Figure 2, this concerns the "N" branches of "Check CQ Completeness" and "Check CQ Consistency", and the "notGranted" branch of "Decide CQ Approval". Next, we separate property checking from directing the control flow. We do this for each node that has more than 1 child. We replace each such node with a process step bearing the same name, followed by an XOR split (BPMN control nodes giving a choice to execution). In the example, this concerns "Check CQ Approval Status". We then re-unite XOR branches using XOR joins (BPMN control nodes leading alternative executions back together), avoiding redundancies in the process by finding pairs of nodes that root identical sub-trees. In Figure 2, this pertains to the two occurrences of "Submit CQ". We introduce a new XOR join node taking the incoming edges of the found node pair. We attach one copy of the common sub-tree below that XOR join. We insert a BPMN start node, join all leaves via a new XOR join, and attach a BPMN end node as the new (only) leaf of the plan. Finally, we introduce parallelism by finding non-interacting sub-sequences of actions in between the XOR splits and joins that were introduced previously. (This is a heuristic notion of parallelism, that does not guarantee to detect all possible parallelizations in the process.)

## 7.2 Positioning of our Prototypes at SAP

As part of the effort to transfer our research results to the SAP product development organization, we integrated our planning approach into two BPM prototypes.

The first one, called *Maestro for BPMN* (Born, Hoffmann, Kaczmarek, Kowalkiewicz, Markovic, Scicluna, Weber, & Zhou, 2008, 2009), is a BPMN process modeling tool developed by SAP Research primarily for the purpose of research and early prototyping. We focus in what follows on our other prototype, which is implemented in the context of the commercial SAP NetWeaver platform (SAP, 2010). NetWeaver is one of the most prominent software platforms at SAP, and is the central platform for SAP's service-oriented architecture. It encompasses all the functionalities required to run an SAP application. Our prototype is implemented as a research extension of *SAP NetWeaver BPM*.

SAP NetWeaver BPM consists of different parts for process modeling and execution. Our planning functionality is integrated into the *SAP NetWeaver BPM Process Composer*, NetWeaver's BPM modeling environment targeted at the creation of new processes. The process modeling is done in BPMN notation; that notation is given an execution semantics by NetWeaver BPM's process execution engine.

## 7.3 Demonstration of our NetWeaver BPM Prototype

We briefly illustrate what using our planning functionality will look like to business users. We consider application scenario (C) as described in Section 3.3, where the business user redesigns a new process from scratch. For application scenarios (A) and (B) from Section 3.3 – using planning during SAM-based development, respectively generating a process template at the beginning of the modeling activity – the same interface can be used.

In designing a new process, the user chooses the "atomic" process steps according to his/her intuition. At IT level, this is no more than drawing a box and inserting a descriptive text. To align this intuitive design with the actual IT infrastructure the process should run on, our planner allows to check how the atomic steps can be implemented based on existing transactions. Say the user has designed the process model shown in Figure 10. Amongst others, the process contains a step "Release Purchase Order", whose intention is to order the purchase of a special part required to satisfy the customer demand, after the customer quote has been accepted. The user now wishes to inflate "Release Purchase Order" to an actual IT-level process fragment having the intended meaning. A double click on the step opens the shown interface for entering the planning initial state and goal, i.e., the desired status variable value changes, associated to this step. The status variable values are chosen via drop-down menus, selecting "initial conditions" on the left-hand side and "goals" on the right-hand side. In the present case, the goal is "PO.Status:Ordered", and the initial condition is "PO.Status:Created" because the purchase order (PO) was already created beforehand and shall now be released.

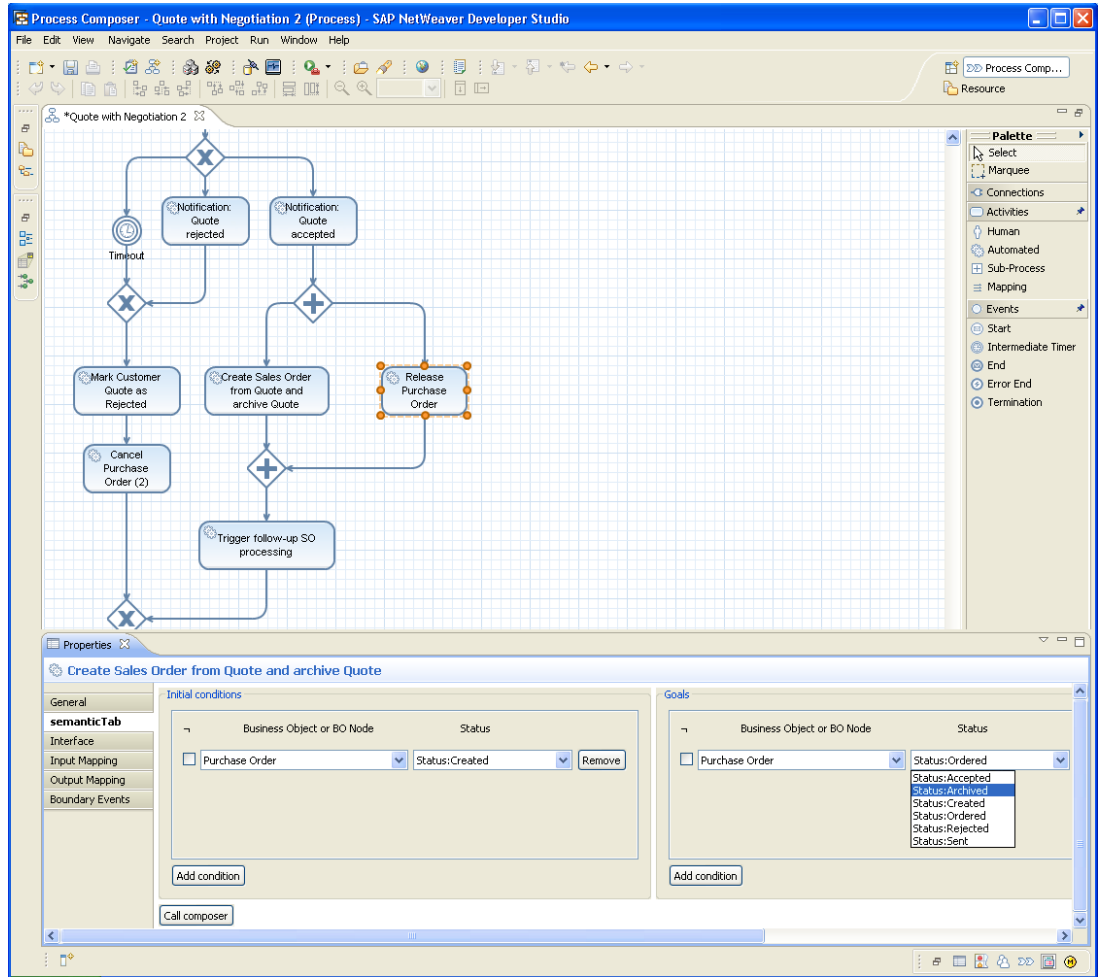

Figure 10: Screenshot of our BPM modeling environment, showing how business users specify planning goals.

Once the status variable values are entered, the user clicks on "Call composer". This invokes the planner, using the specified initial condition/goal to define the SAM planning task.[12] The returned plan is transformed to BPMN, and is inserted into the process model in place of the atomic step that the user had been working on; see the illustration in Figure 11. In the shown case, the plan is a process snippet containing five atomic transactions with one XOR split and two possibly failed outcomes, showing that releasing the PO entails to first process its data, then check this data, and then invoke an approval process similar to that of our illustrative example;[13] finally, the PO is being ordered.

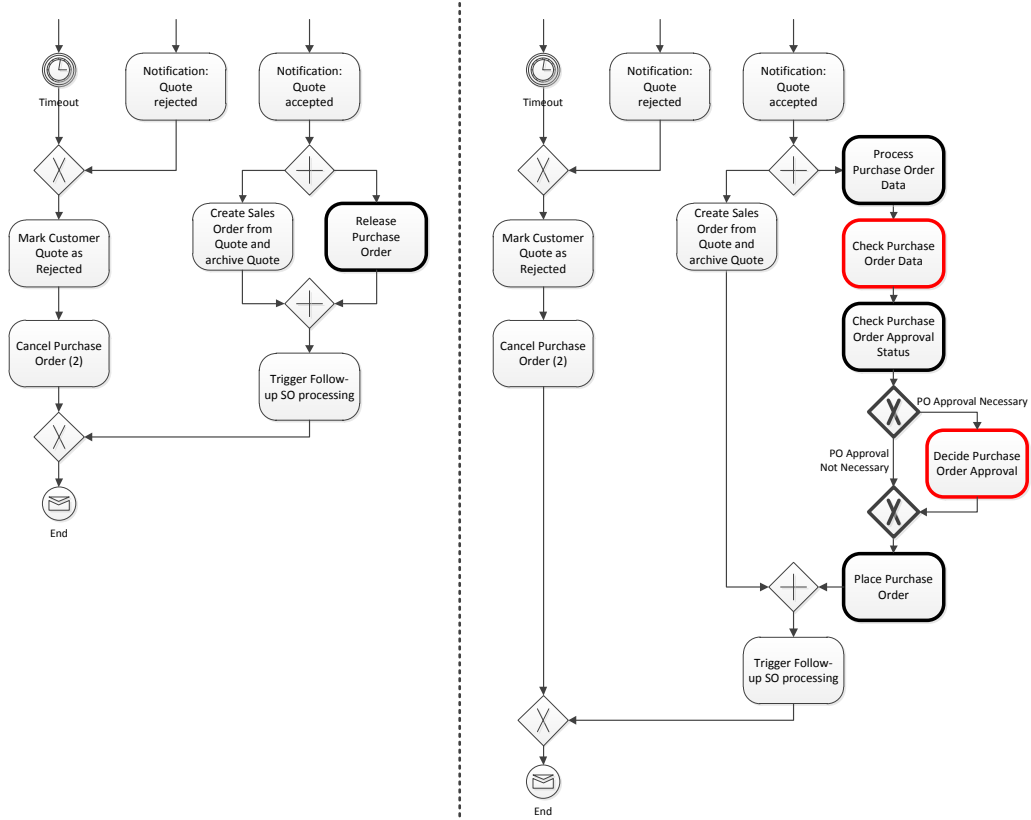

Figure 11: The BPMN process snippet from the screenshot in Figure 10, before (left) and after (right) calling the planner. The affected steps are highlighted in bold; actions with failed outcomes are highlighted in red as before.

Note here that, while cross-BO interactions are not part of the SAM model, the planner helps to create a process that spans multiple BOs, and that indeed ties together functionality that cuts across departmental boundaries. The process snippet shown in Figure 10 invokes the purchase of goods from a supplier as soon as a customer accepts a quote. This is relevant for companies that sell highly customized goods (e.g., special-purpose ships), and who in turn must procure customized parts from their suppliers (e.g., ship engines). Such processing requires to combine services from BOs belonging to Customer Relationship Management (CRM) and Supplier Relationship Management (SRM), and hence from the two "opposite ends" of the system (and company). The designer of the process will typically be intimately familiar with only one of these two, making it especially helpful to be able to call the planner to obtain information about the other one.

---

12. In the current implementation of the prototype, if the value of a variable $x$ is not specified in the "initial condition" given by the user, then the planner does not make use of $x$, i.e., all preconditions on $x$ are assumed to be false until $x$ is set by an action effect. One could of course easily make this more comfortable, by assuming SAM's initial values as a default, and by propagating the effects of earlier SAM transactions (on the same BO) along the process structure. First investigations into the latter have been performed (May & Weber, 2008).

13. Indeed, approval is one of the design patterns that SAP applied throughout SAM. The actual pattern is more complicated than our illustrative version here.

## 7.4 Evaluation of our Prototype at SAP

Our prototype was part of a research transfer project with the NetWeaver BPM group, and shaped during several feedback rounds with developers and architects. The evaluation within SAP consisted mainly of prototype demonstrations at various SAP events. For example, an early version of the tool was demonstrated at the 2008 global meeting of SAP Research "BPM and SI" (SI stands for "Semantic Integration"), which included participants from SAP partners and development. The demonstrations received positive feedback from SAP software architects. The perception was that this functionality would significantly strengthen the link between SAP BPM and the underlying software infrastructure, making it much easier to access the services provided in an effective manner. Most critical comments were focused on some choices in the user interface design, such as "non-logicians will not understand the meaning of the NOT symbol", or "the list of several hundred BOs is too long for a drop-down box".

We do not have customer evaluation data, and it is not foreseeable when we (or anyone else) will be able to obtain such data. When our first prototype became available, a partner organization committed to perform a pilot customer evaluation. That commitment was retracted in the context of the 2008/2009 financial crisis. Anyway, real customer evaluation data may be impossible to come by, let alone publish, due to privacy reasons.

There are also some issues arising from the positioning of our prototype inside the SAP software architecture. The NetWeaver process execution engine currently does not connect to the actual IT services that implement SAM's "actions". While SAM is in productive use within Business ByDesign, NetWeaver BPM is built on a different technology stack. A connection could in principle be established relatively easily – after all, service-orientation is intended to do exactly this sort of thing – however this connection has not as yet been on SAP's agenda, and it involves some SAP-internal political issues. The fact that the main drivers of the presented technology – the authors of this paper – have left the company in the meantime does of course not help to remedy this problem.

## 8. Related Work

The basic idea explored in this paper – using planning systems to help business experts to come up with processes close to the IT infrastructure – has been around for quite a long time. For example, it was mentioned more than a decade ago by Jonathan et al. (1999). It is also discussed in the 2003 roadmap of the PLANET Network of Excellence in AI Planning (Biundo et al., 2003). More recently, Rodriguez-Moreno et al. (2007) implemented the idea in the SHAMASH system. SHAMASH is a knowledge-based BPM tool targeted at helping with the engineering and optimization of process models (Aler, Borrajo, Camacho, & Sierra-Alonso, 2002). The tool includes, amongst other things, user-friendly interfaces allowing users to conveniently annotate processes with rich context information, in particular in the form of "rules" which roughly correspond to planning actions. These rules (and other information) then form the basis for translation into PDDL, and planning for creation of new process models.

The largest body of related work was performed during the last decade under a different name, *semantic web service composition (SWSC)*, in the context of the Semantic Web Community (e.g., Ponnekanti & Fox, 2002; Narayanan & McIlraith, 2002; Srivastava, 2002; Constantinescu, Faltings, & Binder, 2004; Agarwal et al., 2005; Sirin, Parsia, Wu, Hendler, & Nau, 2004; Sirin et al., 2006; Meyer & Weske, 2006; Liu, Ranganathan, & Riabov, 2007). In a nutshell, the idea in SWSC is (1) to annotate web services with some declarative abstract explanation of their functionality, and (2) to exploit these "semantic" annotations to automatically combine web services for achieving a more complex functionality. While SWSC terminology differs from what we use in this paper, the idea is basically the same (although most SWSC works do not address BPM specifically).

The key distinguishing feature of the present work is our approach to obtaining the planning input (the "semantic annotations"). Ours is the first attempt to address the planning/SWSC problem based on SAM, and more generally based on any pre-existing model at all. Since modeling is costly (Kambhampati, 2007; Rodriguez-Moreno et al., 2007), this shift of focus gets us around one of the major open problems in the area. The modeling interfaces in SHAMASH (Rodriguez-Moreno et al., 2007), and some related works attempting to support model creation (Gonzalez-Ferrer, Fernandez-Olivares, & Castillo, 2009; Cresswell, McCluskey, & West, 2009, 2010), address the same problem, but in very different ways and to a less radical extent. Whereas these works attempt to ease the modeling overhead, our re-use of SAM actually removes that overhead completely.

Having said that, of course there are relations between SAM planning and previous work, at the technical level. In particular, the planning and SWSC literature contains a multitude of works dealing with actions that, like SAM's disjunctive effect actions, have more than one possible outcome. Our semantics for such actions, as detailed already in Section 4.2, is a straightforward mixture of two wide-spread notions in planning: observation actions, where one out of a list of possible observations is distinguished, and non-deterministic actions, where one out of a list of possible effects occurs (e.g., Weld et al., 1998; Smith & Weld, 1999; Bonet & Geffner, 2000; Cimatti et al., 2003; Bryce & Kambhampari, 2004; Hoffmann & Brafman, 2005; Bonet & Givan, 2006; Bryce, Kambhampati, & Smith, 2006; Bryce & Buffet, 2008; Palacios & Geffner, 2009).

A prominent line of research in web service composition, known as "the Roman model" (e.g., Berardi, Calvanese, De Giacomo, Lenzerini, & Mecella, 2003, 2005; De Giacomo & Sardiña, 2007; Sardiña, Patrizi, & De Giacomo, 2008; Calvanese, De Giacomo, Lenzerini, Mecella, & Patrizi, 2008), also deals with a notion of "non-determinism" in the component web services, however the framework is very different from ours. The web services in the Roman model are *stateful*. That is, each service has a set of possible own/internal states, and the service provides a set of *operations* to the outside world, which are responsible for transitions in the service's internal state. The composition task is to create a scheduler (a function choosing one service for each operation demanded) interacting with the component services in a way such that they implement a desired goal transition system. Similarly, the web service composition techniques developed by Marco Pistore and his co-workers (e.g.,

Pistore, Marconi, Bertoli, & Traverso, 2005; Bertoli, Pistore, & Traverso, 2006, 2010) deal with this form of non-determinism in stateful component services formalized as transition systems. In their work, the composition task is to create a controller transition system such that the overall (controlled) behavior satisfies a planning-like goal (expressed in the EAGLE language, cf. below). In that latter aspect – attempting to satisfy a planning goal – their framework is slightly closer to ours than the Roman model.

The main distinguishing feature of our formalism is its notion of "weak SAM plans", allowing failed action outcomes but only if they are proved unsolvable, and only if at least one outcome of each action is successful. Some other works have also proposed notions of plans that do not guarantee to achieve the goal in all cases, and some notions of more complex goals can be used to achieve similar effects. We now briefly discuss the notions closest to our own approach.

The notions of weak and strong plans, as discussed in Section 4.2, were first introduced by Cimatti, Giunchiglia, Giunchiglia, and Traverso (1997) and Cimatti, Roveri, and Traverso (1998b), respectively. Strong cyclic plans were first introduced by Cimatti, Roveri, and Traverso (1998a). That notion is orthogonal to weak SAM plans in that neither implies the other. For example, if a "bad" action outcome invalidates, as a side effect, the preconditions of all actions in the task, then that action may form part of a weak SAM plan, but never of a strong cyclic plan. Vice versa, strong cyclic plans have a more general structure, in particular allowing non-deterministic actions to appear more than once in an execution.

Pistore and Traverso (2001) generalize weak, strong, and strong cyclic plans by handling goals taking the form of CTL formulas. However, as pointed out by Dal Lago et al. (2002), such goals are unable to express that *the plan should try to achieve the goal, and give up only if that is not possible.* Dal Lago et al. design the goal language EAGLE which addresses (amongst others) this shortcoming. EAGLE features a variety of goal operators that can be flexibly combined to form goal expressions. One such expression is "TryReach $G_1$ Fail $G_2$", where $G_1$ and $G_2$ are alternative goals. The intuition is that the plan should try to achieve $G_1$, and resort to $G_2$ if reaching $G_1$ has become impossible. More precisely, a plan $T$ for such an EAGLE goal is optimal if, for every state $s$ it traverses: (1) $T$ is a strong plan for $G_1 \vee G_2$; (2) if there exists a strong plan for $G_1$ from $s$, then $T$ is such a strong plan; and (3) if there exists a weak plan for $G_1$ from $s$, then $T$ is such a weak plan. Applying this to our context, say we restrict plans to execute each non-deterministic action at most once. It is easy to see that, within this space of plans, every weak SAM plan is optimal for the EAGLE goal "TryReach $G$ Fail TRUE": weak SAM plans mark $s$ as failed only if reaching $G$ from $s$ is impossible. However, not every action tree $T$ that is optimal for "TryReach $G$ Fail TRUE" is a weak SAM plan. That is because (2) and (3) do not force every action to have at least one solved outcome. In tasks for which no weak SAM plan exists, *any* action tree (e.g., the empty tree) is optimal for "TryReach $G$ Fail TRUE". In tasks with a weak SAM plan, "TryReach $G$ Fail TRUE" forces each solvable action outcome to provide a solution, but imposes no constraints below failed outcomes (which may thus be continued by arbitrarily complex sub-plans).

Shaparau et al. (2006) define a framework that has a similar effect in our context. They consider contingent planning in the presence of a linear preference order over alternative goals. Action trees $T$ are plans if they achieve at least one goal in every leaf, i.e., they are strong plans for the disjunction of the goals. Plan $T$ is at least as good as plan $T'$ if, in every state common to both, the best possible outcome achievable using $T$ is at least as good as that achievable using $T'$. $T$ is optimal if it is at least as good as all other plans. Given this, like for the EAGLE goal "TryReach $G$ Fail TRUE" discussed above, every weak SAM plan is optimal for the goal preference "$G$, TRUE", but the inverse is not true because, in unsolvable tasks and below unsolvable outcomes in solvable tasks, "$G$, TRUE" permits the plan to do anything.

Mediratta and Srivastava (2006) define a framework also based on contingent planning, but where the user provides as additional input a number $K$. Then, (1) a "plan" is any tree $T$ with $\geq K$ leaves achieving the goal; and (2) if such $T$ does not exist, then a "plan" is any tree whose number of such leaves is maximal. Due to (1), failed nodes are not necessarily unsolvable (the plan may simply stop once it reached $K$). Due to (2), even a task where the goal cannot be reached at all, i.e., no matter what the action outcomes are we cannot achieve the goal, has a "plan". In addition, in our application context there is no sensible way, for the human modeler, to choose a meaningful value for $K$.

Summing up, related notions of weak plans exist, but none captures exactly what we want in SAM. On the algorithmic side, all the works listed here use symbolic search (based on BDDs), and are thus quite different from our explicit-state SAM-AO* search. The single exception is the planner described by Mediratta and Srivastava (2006), which is based on a variant of A*, but for a different plan semantics as described.

Research has been performed also into alternative methods, not based on planning, for automatically generating processes. For example, Küster, Ryndina, and Gall (2007) describe a method computing the synchronized product of the life-cycles of a set of business objects, and generating a process corresponding to that product. That process serves as the basis for customer-specific modifications. Clearly, that motivation relates to ours; but the intended meaning of the output (the generated process), and the input assumed for its generation, are quite different. As for the input, Küster et al.'s life-cycles are state machines describing all possible behaviors of the object, which in our formulation here corresponds to the space of all reachable states. That space can be generated based on SAM, but it can be huge even for single BOs, not to mention their product (cf. our results for blind search and scaling across BOs, Sections 6.4 and 6.5). Heuristic search gets us around the need to enumerate all these states. As for the output, Küster et al.'s generated processes guarantee not only to comply with the BO behaviors (which ours do as well), but also to *cover* them, essentially representing all that could be done. This is very different from the plans we generate, whose intention is to show specifically how to move between particular start and end states. Altogether, the methods are complementary. Planning has computational advantages if the involved objects have many possible states, as is often the case in SAM.

## 9. Conclusion

We have pointed out that SAP has built a large-scale model of software behavior, SAM, whose abstraction level and formalization are intimately related to planning models in languages such as PDDL. We have shown how to base a promising BPM application of planning on this fact. Getting the planner input for free, we avoid one of the most important obstacles for making this kind of planning application successful in practice. Our solution is specific to our particular context in its treatment of non-deterministic actions and failed outcomes, but such phenomena are quite common in both planning and web service composition, and our novel approach to dealing with them might turn out to be relevant more generally.

The main open issue is to obtain concrete data evaluating the business value of our application. Some other points are:

- Our modification of FF successfully handles many SAM instances, finding plans within runtimes small enough to apply realistic online-setting cut-offs. About 15% of the instances we encountered still present challenges. These instances could serve as an interesting benchmark for approaches dealing with failed outcomes in ways related to what we do here (cf. Section 8).

- The current SAM model does not reflect dependencies across BOs. Such dependencies do, however, exist in various forms. For example, some BOs form part of the data contained in another kind of BO, some actions on one kind of BO create a new instance of another kind of BO, and some actions must be taken by several BOs together. There is an ongoing activity at SAP Research, aiming at enriching SAM to reflect some of these interactions, with the purpose of more informed model checking. All the interactions can easily be modeled in terms of well-known planning constructs (object creation, and preconditions/effects spanning variables from several BOs), so we expect that this extended model will enable us to generate more accurate plans. As the results from Section 6.5 indicate, additional planning techniques may be required to improve performance in case the number of interacting BOs becomes large. But for smaller numbers (up to around a dozen BOs) the performance of our current tool should still be reasonable.

- SAM currently provides no basis for automatically creating a number of additional process aspects. An important aspect is exception handling, for which at the moment we can only highlight the places (failed nodes) where it needs to be inserted. Another issue is data-flow. This is mostly easy since the application data is already pre-packaged in the relevant BOs, but there are some cases, like security tokens, not covered by this. An open line of research is to determine how these aspects could be modeled, in a way that can be exploited by corresponding planning algorithms.

- It may also be interesting to look into methods presenting the user with a set of alternative processes. For discerning between relevant alternatives, such methods require extensions to SAM, like action duration, action cost, or plan preferences.

From a more general perspective, the key contribution of our work is demonstrating the potential synergy between model-based software engineering and planning-based process generation. Re-using some (or even all) of the required models, the human labor required to realize the planning is dramatically reduced. SAM's methodology – business-level descriptions of individual activities within a software architecture – is not specific to SAP. Thus, exploiting this synergy is a novel approach that may turn out fruitful far beyond the particular application described herein.

## Acknowledgments

We thank the anonymous JAIR reviewers, whose comments helped a lot in improving the paper.

Most of this work was performed while all authors were employed by SAP. Part of this work was performed while Jörg Hoffmann was employed by INRIA (Nancy, France), and while Ingo Weber was employed by The University of New South Wales (Sydney, Australia).

NICTA is funded by the Australian Government as represented by the Department of Broadband, Communications and the Digital Economy and the Australian Research Council through the ICT Centre of Excellence program.

## References

Aalst, W. (1997). Verification of Workflow Nets. In *Application and Theory of Petri Nets 1997*.

Agarwal, V., Chafle, G., Dasgupta, K., Karnik, N., Kumar, A., Mittal, S., & Srivastava, B. (2005). Synthy: A system for end to end composition of web services. *Journal of Web Semantics*, *3*(4).

Aler, R., Borrajo, D., Camacho, D., & Sierra-Alonso, A. (2002). A knowledge-based approach for business process reengineering: SHAMASH. *Knowledge Based Systems*, *15*(8), 473–483.

Bacchus, F. (2000). *Subset of PDDL for the AIPS2000 Planning Competition*. The AIPS-00 Planning Competition Comitee.

Bäckström, C., & Nebel, B. (1995). Complexity results for SAS$^+$ planning. *Computational Intelligence*, *11*(4), 625–655.

Bell, M. (2008). *Service-Oriented Modeling: Service Analysis, Design, and Architecture*. Wiley & Sons.

Berardi, D., Calvanese, D., De Giacomo, G., Lenzerini, M., & Mecella, M. (2003). Automatic composition of e-services that export their behavior. In Orlowska, M. E., Weerawarana, S., Papazoglou, M. P., & Yang, J. (Eds.), *Proceedings of the 1st International Conference on Service-Oriented Computing (ICSOC'03)*, Vol. 2910 of *Lecture Notes in Computer Science*, pp. 43–58. Springer.

Berardi, D., Calvanese, D., De Giacomo, G., Lenzerini, M., & Mecella, M. (2005). Automatic service composition based on behavioral descriptions. *International Journal of Cooperative Information Systems*, *14*(4), 333–376.

Bertoli, P., Pistore, M., & Traverso, P. (2006). Automated web service composition by on-the-fly belief space search. In Long, D., & Smith, S. (Eds.), *Proceedings of the 16th International Conference on Automated Planning and Scheduling (ICAPS-06)*, Ambleside, UK. AAAI.

Bertoli, P., Pistore, M., & Traverso, P. (2010). Automated composition of web services via planning in asynchronous domains. *Artificial Intelligence*, *174*(3-4), 316–361.

Biundo, S., Aylett, R., Beetz, M., Borrajo, D., Cesta, A., Grant, T., McCluskey, L., Milani, A., & Verfaillie, G. (2003). PLANET Technological Roadmap on AI Planning and Scheduling. http://planet.dfki.de/service/Resources/Roadmap/Roadmap2.pdf.

Bonet, B., & Geffner, H. (2000). Planning with incomplete information as heuristic search in belief space. In Chien, S., Kambhampati, R., & Knoblock, C. (Eds.), *Proceedings of the*

*5th International Conference on Artificial Intelligence Planning Systems (AIPS-00)*, pp. 52–61, Breckenridge, CO. AAAI Press, Menlo Park.

Bonet, B., & Geffner, H. (2001). Planning as heuristic search. *Artificial Intelligence*, *129*(1–2), 5–33.

Bonet, B., & Givan, B. (2006). 5th international planning competition: Non-deterministic track – call for participation. In *Proceedings of the 5th International Planning Competition (IPC'06)*.

Born, M., Hoffmann, J., Kaczmarek, T., Kowalkiewicz, M., Markovic, I., Scicluna, J., Weber, I., & Zhou, X. (2008). Semantic annotation and composition of business processes with Maestro. In *Demonstrations at ESWC'08: 5th European Semantic Web Conference*, pp. 772–776, Tenerife, Spain.

Born, M., Hoffmann, J., Kaczmarek, T., Kowalkiewicz, M., Markovic, I., Scicluna, J., Weber, I., & Zhou, X. (2009). Supporting execution-level business process modeling with semantic technologies. In *Demonstrations at DASFAA'09: Database Systems for Advanced Applications*, pp. 759–763, Brisbane, Australia.

Bryce, D., & Buffet, O. (2008). 6th international planning competition: Uncertainty part. In *Proceedings of the 6th International Planning Competition (IPC'08)*.

Bryce, D., & Kambhampari, S. (2004). Heuristic guidance measures for conformant planning. In Koenig, S., Zilberstein, S., & Koehler, J. (Eds.), *Proceedings of the 14th International Conference on Automated Planning and Scheduling (ICAPS-04)*, pp. 365–374, Whistler, Canada. AAAI.

Bryce, D., Kambhampati, S., & Smith, D. E. (2006). Planning graph heuristics for belief space search. *Journal of Artificial Intelligence Research*, *26*, 35–99.

Bylander, T. (1994). The computational complexity of propositional STRIPS planning. *Artificial Intelligence*, *69*(1–2), 165–204.

Calvanese, D., De Giacomo, G., Lenzerini, M., Mecella, M., & Patrizi, F. (2008). Automatic service composition and synthesis: the roman model. *IEEE Data Engineering Bulletin*, *31*(3), 18–22.

Cimatti, A., Giunchiglia, F., Giunchiglia, E., & Traverso, P. (1997). Planning via model checking: A decision procedure for ar. In Steel, S., & Alami, R. (Eds.), *Recent Advances in AI Planning. 4th European Conference on Planning (ECP'97)*, Vol. 1348 of *Lecture Notes in Artificial Intelligence*, pp. 130–142, Toulouse, France. Springer-Verlag.

Cimatti, A., Pistore, M., Roveri, M., & Traverso, P. (2003). Weak, strong, and strong cyclic planning via symbolic model checking. *Artificial Intelligence*, *147*(1-2), 35–84.

Cimatti, A., Roveri, M., & Traverso, P. (1998a). Automatic obdd-based generation of universal plans in non-deterministic domains. In Mostow, J., & Rich, C. (Eds.), *Proceedings of the 15th National Conference of the American Association for Artificial Intelligence (AAAI-98)*, pp. 875–881, Madison, WI, USA. MIT Press.

Cimatti, A., Roveri, M., & Traverso, P. (1998b). Strong planning in non-deterministic domains via model checking. In Simmons, R., Veloso, M., & Smith, S. (Eds.), *Proceedings of the 4th International Conference on Artificial Intelligence Planning Systems (AIPS-98)*, pp. 36–43, Pittsburgh, PA. AAAI Press, Menlo Park.

Cohn, D., & Hull, R. (2009). Business artifacts: A data-centric approach to modeling business operations and processes. *IEEE Data Engineering Bulletin*, 3–9.

Constantinescu, I., Faltings, B., & Binder, W. (2004). Large scale, type-compatible service composition. In Jain, H., & Liu, L. (Eds.), *Proceedings of the 2nd International Conference on Web Services (ICWS-04)*, pp. 506–513, San Diego, California, USA. IEEE Computer Society.

Cresswell, S., McCluskey, T., & West, M. (2010). Acquiring planning domains models using LOCM. *The Knowledge Engineering Review*.

Cresswell, S., McCluskey, T. L., & West, M. M. (2009). Acquisition of object-centred domain models from planning examples. In Gerevini, A., Howe, A. E., Cesta, A., & Refanidis, I. (Eds.), *Proceedings of the 19th International Conference on Automated Planning and Scheduling (ICAPS-09)*, Sydney, Australia. AAAI.

Dal Lago, U., Pistore, M., & Traverso, P. (2002). Planning with a language for extended goals. In Dechter, R., Kearns, M., & Sutton, R. (Eds.), *Proceedings of the 18th National Conference of the American Association for Artificial Intelligence (AAAI-02)*, pp. 447–454, Edmonton, AL, USA. MIT Press.

De Giacomo, G., & Sardiña, S. (2007). Automatic synthesis of new behaviors from a library of available behaviors. In Veloso, M. (Ed.), *Proceedings of the 20th International Joint Conference on Artificial Intelligence (IJCAI-07)*, pp. 1866–1871, Hyderabad, India. Morgan Kaufmann.

Dumas, M., ter Hofstede, A., & van der Aalst, W. (Eds.). (2005). *Process Aware Information Systems: Bridging People and Software Through Process Technology*. Wiley Publishing.

Fox, M., & Long, D. (2003). PDDL2.1: An extension to PDDL for expressing temporal planning domains. *Journal of Artificial Intelligence Research*, *20*, 61–124.

Gazen, B. C., & Knoblock, C. (1997). Combining the expressiveness of UCPOP with the efficiency of Graphplan. In Steel, S., & Alami, R. (Eds.), *Recent Advances in AI Planning. 4th European Conference on Planning (ECP'97)*, Vol. 1348 of *Lecture Notes in Artificial Intelligence*, pp. 221–233, Toulouse, France. Springer-Verlag.

Gerevini, A., Haslum, P., Long, D., Saetti, A., & Dimopoulos, Y. (2009). Deterministic planning in the fifth international planning competition: PDDL3 and experimental evaluation of the planners. *Artificial Intelligence*, *173*(5-6), 619–668.

Gonzalez-Ferrer, A., Fernandez-Olivares, J., & Castillo, L. (2009). JABBAH: a Java application framework for the translation between business process models and HTN. In *Proceedings of the 3rd International Competition on Knowledge Engineering for Planning and Scheduling*, Thessaloniki, Greece.

Helmert, M. (2006). The Fast Downward planning system. *Journal of Artificial Intelligence Research*, *26*, 191–246.

Helmert, M. (2009). Concise finite-domain representations for pddl planning tasks. *Artificial Intelligence*, *173*(5-6), 503–535.

Hoffmann, J., & Brafman, R. (2005). Contingent planning via heuristic forward search with implicit belief states. In Biundo, S., Myers, K., & Rajan, K. (Eds.), *Proceedings of the 15th International Conference on Automated Planning and Scheduling (ICAPS-05)*, pp. 71–80, Monterey, CA, USA. AAAI.

Hoffmann, J., & Edelkamp, S. (2005). The deterministic part of IPC-4: An overview. *Journal of Artificial Intelligence Research, 24*, 519–579.

Hoffmann, J., & Nebel, B. (2001). The FF planning system: Fast plan generation through heuristic search. *Journal of Artificial Intelligence Research, 14*, 253–302.

Jonathan, P. J., Moore, J., Stader, J., Macintosh, A., & Chung, P. (1999). Exploiting ai technologies to realise adaptive workflow systems. In *Proceedings ot the AAAI'99 Workshop on Agent-Based Systems in the Business Context*.

Kambhampati, S. (2007). Model-lite planning for the web age masses: The challenges of planning with incomplete and evolving domain models. In Howe, A., & Holte, R. C. (Eds.), *Proceedings of the 22nd National Conference of the American Association for Artificial Intelligence (AAAI-07)*, Vancouver, BC, Canada. MIT Press.

Kitchin, D. E., McCluskey, T. L., & West, M. M. (2005). B vs ocl: Comparing specification languages for planning domains. In *Proceedings of the ICAPS'05 Workshop on Verification and Validation of Model-Based Planning and Scheduling Systems*.

Krafzig, D., Banke, K., & Slama, D. (2005). *Enterprise SOA: Service-Oriented Architecture Best Practices*. Prentice Hall.

Küster, J. M., Ryndina, K., & Gall, H. (2007). Generation of business process models for object life cycle compliance. In Alonso, G., Dadam, P., & Rosemann, M. (Eds.), *Proceedings of the 5th International Conference on Business Process Management (BPM'07)*, Vol. 4714 of *Lecture Notes in Computer Science*, pp. 165–181. Springer.

Liu, Z., Ranganathan, A., & Riabov, A. (2007). A planning approach for message-oriented semantic web service composition. In Howe, A., & Holte, R. C. (Eds.), *Proceedings of the 22nd National Conference of the American Association for Artificial Intelligence (AAAI-07)*, Vancouver, BC, Canada. MIT Press.

May, N., & Weber, I. (2008). Information gathering for semantic service discovery and composition in business process modeling. In *CIAO!'08: Workshop on Cooperation & Interoperability - Architecture & Ontology at CAiSE'08*, Vol. LNBIP 10, pp. 46–60, Montpellier, France.

McDermott, D., Ghallab, M., Howe, A., Knoblock, C., Ram, A., Veloso, M., Weld, D., & Wilkins, D. (1998). PDDL – the planning domain definition language. Tech. rep. CVC TR-98-003, Yale Center for Computational Vision and Control.

McDermott, D. V. (1999). Using regression-match graphs to control search in planning. *Artificial Intelligence, 109*(1-2), 111–159.

Mediratta, A., & Srivastava, B. (2006). Applying planning in composition of web services with a user-driven contingent planner. IBM Research Report RI 06002.

Meyer, H., & Weske, M. (2006). Automated service composition using heuristic search. In Dustdar, S., Fiadeiro, J. L., & Sheth, A. P. (Eds.), *Proceedings of the 4th International*

*Conference on Business Process Management (BPM'06)*, Vol. 4102 of *Lecture Notes in Computer Science*, pp. 81–96. Springer.

Narayanan, S., & McIlraith, S. (2002). Simulation, verification and automated composition of web services. In Iyengar, A., & Roure, D. D. (Eds.), *Proceedings of the 11th International World Wide Web Conference (WWW-02)*, Honolulu, Hawaii, USA. ACM.

Nilsson, N. J. (1969). Searching problem-solving and game-playing trees for minimal cost solutions. In *Information Processing 68 Vol. 2*, pp. 1556–1562, Amsterdam, Netherlands.

Nilsson, N. J. (1971). *Problem Solving Methods in Artificial Intelligence*. McGraw-Hill.

Object Management Group (2006). Object Constraint Language Specification, Version 2. http://www.omg.org/technology/documents/formal/ocl.htm.

Object Management Group (2008). Business Process Modeling Notation, V1.1. http://www.bpmn.org/.

Palacios, H., & Geffner, H. (2009). Compiling uncertainty away in conformant planning problems with bounded width. *Journal of Artificial Intelligence Research*, *35*, 623–675.

Pearl, J. (1984). *Heuristics*. Morgan Kaufmann.

Pednault, E. P. (1989). ADL: Exploring the middle ground between STRIPS and the situation calculus. In Brachman, R., Levesque, H. J., & Reiter, R. (Eds.), *Principles of Knowledge Representation and Reasoning: Proceedings of the 1st International Conference (KR-89)*, pp. 324–331, Toronto, ON. Morgan Kaufmann.

Pesic, M., Schonenberg, M. H., Sidorova, N., & van der Aalst, W. M. P. (2007). Constraint-based workflow models: Change made easy. In Meersman, R., & Tari, Z. (Eds.), *OTM Conferences (1)*, Vol. 4803 of *Lecture Notes in Computer Science*, pp. 77–94. Springer.

Pistore, M., Marconi, A., Bertoli, P., & Traverso, P. (2005). Automated composition of web services by planning at the knowledge level. In Kaelbling, L. (Ed.), *Proceedings of the 19th International Joint Conference on Artificial Intelligence (IJCAI-05)*, Edinburgh, Scotland. Morgan Kaufmann.

Pistore, M., & Traverso, P. (2001). Planning as model checking for extended goals in non-deterministic domains. In Nebel, B. (Ed.), *Proceedings of the 17th International Joint Conference on Artificial Intelligence (IJCAI-01)*, pp. 479–486, Seattle, Washington, USA. Morgan Kaufmann.

Ponnekanti, S., & Fox, A. (2002). SWORD: A developer toolkit for web services composition. In Iyengar, A., & Roure, D. D. (Eds.), *Proceedings of the 11th International World Wide Web Conference (WWW-02)*, Honolulu, Hawaii, USA. ACM.

Richter, S., & Helmert, M. (2009). Preferred operators and deferred evaluation in satisficing planning. In Gerevini, A., Howe, A. E., Cesta, A., & Refanidis, I. (Eds.), *Proceedings of the 19th International Conference on Automated Planning and Scheduling (ICAPS-09)*, Sydney, Australia. AAAI.

Rodriguez-Moreno, M. D., Borrajo, D., Cesta, A., & Oddi, A. (2007). Integrating planning and scheduling in workflow domains. *Expert Systems Applications*, *33*(2), 389–406.

SAP (2010). SAP NetWeaver.. http://www.sap.com/platform/netweaver/index.epx.

Sardiña, S., Patrizi, F., & De Giacomo, G. (2008). Behavior composition in the presence of failure. In Brewka, G., & Lang, J. (Eds.), *Proceedings of the 11th International Conference on Principles of Knowledge Representation and Reasoning (KR'08)*, pp. 640–650. AAAI Press.

Schneider, S. (2001). *The B-Method: An Introduction.* Palgrave.

Shaparau, D., Pistore, M., & Traverso, P. (2006). Contingent planning with goal preferences. In Gil, Y., & Mooney, R. J. (Eds.), *Proceedings of the 21st National Conference of the American Association for Artificial Intelligence (AAAI-06)*, Boston, Massachusetts, USA. MIT Press.

Sirin, E., Parsia, B., Wu, D., Hendler, J., & Nau, D. (2004). HTN planning for web service composition using SHOP2. *Journal of Web Semantics, 1*(4).

Sirin, E., Parsia, B., & Hendler, J. (2006). Template-based composition of semantic web services. In *AAAI Fall Symposium on Agents and Search.*

Smith, D. E., & Weld, D. S. (1999). Temporal planning with mutual exclusion reasoning. In Dean, T. (Ed.), *Proceedings of the 16th International Joint Conference on Artificial Intelligence (IJCAI-99)*, pp. 326–337, Stockholm, Sweden. Morgan Kaufmann.

Srivastava, B. (2002). Automatic web services composition using planning. In *Knowledge Based Computer Systems (KBCS-02)*, pp. 467–477.

Traverso, P., Ghallab, M., & Nau, D. (Eds.). (2005). *Automated Planning: Theory and Practice.* Morgan Kaufmann.

Turner, J., & McCluskey, T. L. (1994). *The Construction of Formal Specifications: an Introduction to the Model-Based and Algebraic Approaches.* McGraw Hill Software Engineering series.

van der Aalst, W. (2003). Business process management demystified: A tutorial on models, systems and standards for workflow management. In *Lectures on Concurrency and Petri Nets in ACPN'04: Advanced Courses in Petri Nets*, pp. 1–65.

van der Aalst, W. M. P., & Pesic, M. (2006). Decserflow: Towards a truly declarative service flow language. In Bravetti, M., Núñez, M., & Zavattaro, G. (Eds.), *WS-FM*, Vol. 4184 of *Lecture Notes in Computer Science*, pp. 1–23. Springer.

Wainer, J., & de Lima Bezerra, F. (2003). *Groupware: Design, Implementation, and Use*, Vol. 2806 of *LNCS*, chap. Constraint-based flexible workflows, pp. 151–158. Springer-Verlag.

Weld, D. S., Anderson, C. R., & Smith, D. E. (1998). Extending graphplan to handle uncertainty & sensing actions. In Mostow, J., & Rich, C. (Eds.), *Proceedings of the 15th National Conference of the American Association for Artificial Intelligence (AAAI-98)*, pp. 897–904, Madison, WI, USA. MIT Press.

Weske, M. (2007). *Business Process Management: Concepts, Languages, Architectures.* Springer-Verlag.

Yoon, S. W., Fern, A., & Givan, R. (2007). FF-Replan: A baseline for probabilistic planning. In Boddy, M., Fox, M., & Thiebaux, S. (Eds.), *Proceedings of the 17th International Conference on Automated Planning and Scheduling (ICAPS-07)*, Providence, Rhode Island, USA. AAAI.

Younes, H., Littman, M., Weissman, D., & Asmuth, J. (2005). The first probabilistic track of the international planning competition. *Journal of Artificial Intelligence Research*, *24*, 851–887.